\pdfoutput=1
\documentclass[conference]{IEEEtran}
\IEEEoverridecommandlockouts
\usepackage{booktabs} % For \setcopyright{rightsretained}
\usepackage{balance}
\usepackage{amsmath,amssymb,amsfonts}
\usepackage{graphicx}
\usepackage{textcomp}
\usepackage{xcolor}
\usepackage{amsmath}
\usepackage{float}
\usepackage{amsfonts}
\usepackage{bbm}
\usepackage{bbm}
\usepackage{graphicx}
\usepackage{varwidth}
\usepackage{caption}
\usepackage{enumitem}
\usepackage{dsfont}
\usepackage{tabularx}
\usepackage{booktabs}
\usepackage{subfigure}
\usepackage{pgfplots}
\pgfplotsset{compat=1.8}
\usepgfplotslibrary{statistics}

\usepackage[acronym, nomain,nonumberlist]{glossaries}

\makeatletter
\newif\if@restonecol
\makeatother

\usepackage[ruled,vlined,linesnumbered]{algorithm2e}

\usepackage[textsize=tiny, textwidth=1.5cm]{todonotes}

\usepackage[normalem]{ulem}
\usepackage{acro}

\DeclareAcronym{ems}{
  short = EMS,
  long  = Emergency Medical Services,
  sort  = E,
}
\DeclareAcronym{mdp}{
  short = MDP,
  long  = Markov Decision Process,
  sort  = M,
}
\DeclareAcronym{smdp}{
  short = SMDP,
  long  = Semi-Markov Decision Process,
  sort  = S,
}

\DeclareAcronym{dtmdp}{
  short = DTMDP,
  long  = Discrete Time Markov Decision Process,
  sort  = D,
}

\DeclareAcronym{lstm}{
  short = LSTM,
  long  = Long Short-Term Memory Neural Network,
  sort  = L,
}
\DeclareAcronym{aft}{
  short = AFT,
  long  = Accelerated Failure Model,
  sort  = A,
}
\DeclareAcronym{mle}{
  short = MLE,
  long  = Maximum Likelihood Estimation,
  sort  = M,
}

\DeclareAcronym{mcts}{
  short = MCTS,
  long  = Monte Carlo Tree Search,
  sort  = M,
}

\DeclareAcronym{alt}{
  short = ALT,
  long  = A* Search with Landmarks,
  sort  = A,
}
\DeclareAcronym{hcps}{
  short = HCPS,
  long  = Human-in-the-Loop Cyber-Physical Systems,
  sort  = H,
}

\begin{document}

% \copyrightyear{2019} 
% \acmYear{2019} 
% \setcopyright{acmcopyright}
% \acmConference[ICCPS '19]{10th ACM/IEEE International Conference on Cyber-Physical Systems (with CPS-IoT Week 2019)}{April 16--18, 2019}{Montreal, QC, Canada}
% \acmBooktitle{10th ACM/IEEE International Conference on Cyber-Physical Systems (with CPS-IoT Week 2019) (ICCPS '19), April 16--18, 2019, Montreal, QC, Canada}
% \acmPrice{15.00}
% \acmDOI{10.1145/3302509.3311055}
% \acmISBN{978-1-4503-6285-6/19/04}

\title{An Online Decision-Theoretic Pipeline for Responder Dispatch
\thanks{* These authors had equal contribution in this work}
}

\author{\IEEEauthorblockN{Ayan Mukhopadhyay*}
\IEEEauthorblockA{\textit{Vanderbilt University}\\
Nashville, TN, USA}
\and
\IEEEauthorblockN{Geoffrey Pettet*}
\IEEEauthorblockA{\textit{Vanderbilt University}\\
Nashville, TN, USA}
\and
\IEEEauthorblockN{Chinmaya Samal}
\IEEEauthorblockA{\textit{Vanderbilt University}\\
Nashville, TN, USA}
\and
\IEEEauthorblockN{Abhishek Dubey}
\IEEEauthorblockA{\textit{Vanderbilt University}\\
Nashville, TN, USA}
\and
\IEEEauthorblockN{Yevgeniy Vorobeychik}
\IEEEauthorblockA{\textit{Washington University}\\
St. Louis, MO, USA}
}

\maketitle

\begin{abstract}
The problem of dispatching emergency responders to service traffic accidents, fire, distress calls and crimes plagues urban areas across the globe. While such problems have been extensively looked at, most approaches are offline. Such methodologies fail to capture the dynamically changing environments under which critical emergency response occurs, and therefore, fail to be implemented in practice. Any holistic approach towards creating a pipeline for effective emergency response must also look at other challenges that it subsumes - predicting when and where incidents happen and understanding the changing environmental dynamics. We describe a system that collectively deals with all these problems in an online manner, meaning that the models get updated with streaming data sources. We highlight why such an approach is crucial to the effectiveness of emergency response, and present an algorithmic framework that can compute promising actions for a given decision-theoretic model for responder dispatch. We argue that carefully crafted heuristic measures can balance the trade-off between computational time and the quality of solutions achieved and highlight why such an approach is more scalable and tractable than traditional approaches. We also present an online mechanism for incident prediction, as well as an approach based on recurrent neural networks for learning and predicting environmental features that affect responder dispatch. We compare our methodology with prior state-of-the-art and existing dispatch strategies in the field, which show that our approach results in a reduction in response time with a drastic reduction in computational time.
%\ad{check if this last statement is true}
%\ad{Ayan please mention heuristics and tradeoff between the optimality of the result and the computation time. Effectively the result of this paper is going to be an approximation of the best answer}
\end{abstract}
% {\footnotesize
% \glsaddall
% \printglossary[type=\acronymtype,title=Acronyms]
% }

\acuseall

{\printacronyms
}
\vspace{-0.1in}
\section{Introduction} \label{sec:intro}

\textbf{Emerging Trends and Challenges}:
Smart and connected communities are Human-in-the-Loop Cyber-Physical systems (H-CPS), with interactions between humans, the outside environment, and computational tools that assist in decision-making processes \cite{cassandras2016smart}. Analysis and optimization of H-CPS's is challenging primarily due to the inherent complexity and the sheer number of agents involved. Making accurate models is difficult, and simple rule based strategies often fail to capture the dynamics of the problem space.

Consider the classical problem of emergency response. The goal of responders is to minimize the variance in the operational delay between the time incidents are reported and when responders arrive on the scene. However, solving this problem requires not just sending the nearest emergency responder, but sometimes being proactive placing emergency vehicles in regions with higher incident likelihood. Sending the nearest available responder by euclidean distance ignores road networks and their congestion, as well as where the resources are stationed. Greedily assigning resources to incidents can lead to resources being pulled away from their stations, increasing response times if an incident occurs in the future in the area where responder should be positioned. 

%Advancements in sensors and middleware provide opportunities to collect large amounts of data on complex operations and processes of H-CPS's, which can be used to develop data driven operational techniques. 

%{\bf Challenges: } Analysis and optimization of H-CPS's is challenging. Due to the inherent complexity and the sheer number of agents involved, making accurate models is difficult and simple rule based strategies often fail to capture the dynamics of the problem space. Nashville Fire Department's current strategy when incidents occur, for example, is to send the nearest available responder by euclidean distance. This strategy ignores road networks and their congestion, as well as where the resources are stationed. Resources can be pulled away from their stations, increasing response times if an incident occurs in the future where they should be positioned. 

Data-driven approaches have been shown to produce more informed solutions to such problems \cite{dataDrivenSmart} -- examples include predicting crime and traffic accidents in urban areas \cite{mukhopadhyayGameSec16,mukhopadhyayAAMAS17}, and
%including  design of 
building architectures for smart city ecosystems \cite{abu2017data}. In this paper, we leverage the potential of data-driven approaches and utilize real-world incident data for making informed decisions about effective stationing and dispatch of Emergency Medical Services (EMS) resources in a large urban community (Nashville, TN).

{\bf Contributions:} We break down the problem of responder dispatch into three atomic sub-components: incident prediction, environment simulation, and the dispatching approach. Our contributions are as follows:

\begin{itemize}
    \item \underline{Incident Prediction: Online Survival Analysis} - We define a novel online approach to incident prediction that predicts incidents in time and space. Previous work in this domain has treated this as a batch learning problem \cite{aamas16,mukhopadhyayAAMAS17,mukhopadhyayGameSec16}, in which incident prediction models are learned once, and are subsequently used to aid response decisions. This fails to capture the changing dynamics of urban systems in which emergency responders operate, and we bridge this gap by creating an online incident prediction algorithm.
%    \item This is crucial to making dispatch decisions - in order to effectively allocate and dispatch responders, one must first understand where and when incidents happen. In order to take into account the latest spatial and temporal incident patterns, we propose an online learning algorithm based on survival analysis.  

    \item \underline{Dispatch Algorithm} - We formulate the problem of dispatching responders to incidents as a Semi Markov Decision Process (SMDP). Such an approach has recently been shown to work exceptionally well in this domain \cite{mukhopadhyayAAMAS18}. However, such systems have enormous computational load that limit their deployment in practice. We highlight this issue through the course of the paper and design an efficient solution that is fast, scalable and can work in a dynamic environment.
%    \item captures changing environmental factors as they occur and quickly computes near-optimal decisions to aid responders.

    \item \underline{Decision Theoretic Framework} - We compose the Incident Prediction, Environment Simulation, and Dispatching components into a framework that makes real time dispatching decisions based on traffic congestion and predicted incident distributions. Each component is modular, so improvements are easy to integrate into the framework. 

\end{itemize}

{\bf Outline:} We present and evaluate each of the components separately, as well as the entire system that combines them into an online pipeline and show that it results in better performance, and a remarkable decrease in computational run-time. We begin by presenting a high-level system model and problem description in Section \ref{Sec:SystemModel}, and present our solution in Section \ref{sec:solution}. We show our empirical evaluation in Section \ref{Sec:Experiments}, go over a summary of prior work in the field in Section \ref{Sec:PriorWork} and summarize the paper in Section \ref{Sec:Conclusion}. Table \ref{tab:notation} describes the symbols used.

\begin{table}[t]
\footnotesize
\centering
\caption{Notation Table}
\label{tab:notation}
\begin{tabular}{|c|c|}
\hline
Symbol & Meaning                                      \\ \hline
$G$    & Set of equally sized grids                   \\ \hline
$R$    & Set of Responders                            \\ \hline
$t$    & Arrival-time between incidents               \\ \hline
$w$    & Features that affect incident arrival        \\ \hline
$f$    & A distribution over $t$, conditional on $w$  \\ \hline
$M_s$  & Responder Dispatch SMDP                      \\ \hline
$M_d$  & Responder Dispatch Discrete-Time MDP (DTMDP) \\ \hline
$h$    & Horizon of the Monte-Carlo Search Tree       \\ \hline
$D$    & Historical Dataset of incidents              \\ \hline
$D^{'}$ & Stream Dataset of incidents              \\ 
\hline
$L$    & Log-Likelihood of Incidents                  \\ \hline
$\Theta$ & A Generative Model of the Urban Area \\
\hline
\end{tabular}
\vspace{-0.1in}
\end{table}

\section{Problem Description}\label{Sec:SystemModel}

The problem deals with an urban area, in which incidents like traffic accidents, fires, distress calls and crimes happen in space and time. Such incidents are reported to a central emergency response system, which then dispatches responders like police vehicles and ambulances. This system governs the entire pipeline of incident response, including detecting and reporting incidents, monitoring and placing a fleet of response vehicles, and finally dispatching responders when incidents occur. Such responders are equipped with devices that facilitate communication to and from central control stations. They are then dispatched by a human (guided by some algorithmic approach), a process which typically takes seconds, but can be longer if dispatchers are busy \cite{fireDepartmentCommunication}. 

For simplicity we discuss our approach with a single responder type and a single type of incident, but such \textit{homogeneity} is not required for this approach. 

Formally, we consider that the entire urban area is divided into as set of grids $G$. Incidents happen in these grids with an inter-arrival temporal distribution $f$, conditional on a set of features $w$. Such incidents need to be responded to by a set of responders $R$. Each responder is allocated to a specific \textit{depot}, which are immobile stations located in a particular grid. Once a responder has finished servicing an incident, it is directed back to its depot and becomes available to be re-dispatched while in route. We also assume that if there are any free responders when an incident is reported, then some responder must be dispatched to attend to the incident. This is a direct consequence of the legal bounds within which emergency responders operate, as well as of the critical nature of the incidents. If an incident happens and there are no free responders available, then the incident enters a non-priority waiting queue and is attended to when responders become free.

Dispatch Systems in use today by major metropolitan areas such as Nashville work as follows: when an incident is reported, the system dispatches the closest available responder using the euclidean distance (i.e. "as the crow flies") between the incident and the responder's current position \cite{fireDepartmentCommunication}. This method has several disadvantages: 1) The euclidean distance between two locations is not necessarily representative of the actual time to travel between them since travel time depends on the road network and its current congestion. 2) By using the responder's current location, it ignores where the responder should be stationed. Nashville`s incident response data shows that this can lead to responders being pulled away from their depots, causing future incidents around those depots to have longer response times. 3) This method ignores the likely future incident distribution. Although dispatching the closest responder is greedily optimal, it may not be the best choice given the distribution of future incidents.

This motivates us to model responder dispatch formally. We begin by introducing the problem formulation of the online dispatch system in this section. We denote by $\tau \sim f$, a random variable that represents time between incidents in the urban area. 
%First, we denote by $t$ a random variable that represents the time between incidents in the urban area. Then, we assume that a probability distribution $f(t|w)$ governs the occurrence of incidents, with $w$ being a set of arbitrary features (discussed in section \ref{sec:incidentPrediction}). We make the natural assumption that all the factors that determine incident arrival are captured in the feature space.
% Also, we consider a homogeneous model over incident arrival for all grids. The differences in rates of incidents in different grids is captured in the feature space of the model. 

Given such a model of incident arrival, we now look at the model for responder dispatch. We formally model the problem of dynamic incident response as a semi-Markov decision process (SMDP)~\cite{mdpHu,mukhopadhyayAAMAS18}, and refer to this process as $M_s$.
An SMDP is described by the following tuple,

\vspace{-0.2in}
\begin{equation}\label{eq:SMDP}
\begin{aligned}
    \{S,A,p_{ij}(a),t(i,j,a),\rho(i,a) ,\alpha\}
\end{aligned}
\end{equation}
where $S$ is a finite state space,  $A$ is the set of actions, $p_{ij}(a)$ is the probability with which the process transitions from state $i$ to state $j$ when action $a$ is taken, $t(i,j,a)$ is a distribution over the time spent during the transition from state $i$ to state $j$ under action $a$, $\rho(i,a)$ is the reward received when action $a$ is taken in state $s_i$, and $\alpha$ is the discount factor for future rewards. 

\textbf{States:}
At any point in time $t$, the state of the problem consists of the a tuple $\{I^t,R^t,E^t\}$, where $I^t$ is a collection of grid indices that are waiting to be serviced, ordered according their times of occurrence. $R^t$ represents a collection of vectors, where $r^t_i \in R^t$ captures all relevant information about the $i^\text{th}$ responder such as it's current position and status.
% Formally, $r^t_i \in R^t$ is a set $\{h^t_i,p^t_i,d^t_i,c^t_i\}$, where $h^t_i$ corresponds to the depot that responder $i$ is assigned to, $p^t_i$ is the position of responder $i$, $d^t_i$ is the destination that it is traveling to (where $d^t_i = 0$ indicates that responder $i$ has no destination assigned), and $c^t_i$ is its current condition, all observed at the state of our world at time $t$. Observe that $h^t_i,p^t_i,d^{t}_i \in G$ $\forall t, i$.
The state variable $E^t$ captures relevant environmental factors that affect dynamic dispatch of responders at time $t$. Such factors are problem specific, so we leave the choice of such features to the designer of the responder system, but describe the specific features used in our system later. 
%We point out that we overload the notation for states for convenience, and use state $s_i$ to refer to an arbitrary state that is identified by the index $i$. 
% This lets us simplify notation by not explicitly referring to time when it is not needed.

\textbf{Actions:}
Actions in our world correspond to directing responders either to incidents or back to their depots, when the process encounters a decision-making state. 
% We point out that actions can not be taken at all states, as only a subset of states present the scope of decision-making. 
We denote the set of all actions by $A$
% , with an action by $a^{i}_{kj} \in A$ corresponding to a decision to send a responder which is from depot $k$ and is currently in grid $j$, to an incident at grid $i$.
and refer to a generic action by $a$. We use $A(s^i)$ to denote the set of actions that are available in state $s^i \in S$, and impose a constraint that whenever at least one responder is available and an incident occurs, we immediately dispatch \emph{some responder}. Since the entire process evolves in continuous time, one can consider the existence of a single decision-making state at any instant, which leads to a single action being needed.

%Since the entire process evolves in continuous time, we can always consider that a single decision-making state exists at any point in time. A direct consequence of this is that at any point in time, a single action needs to be taken in our model. 

\textbf{Transitions:}
The SMDP model evolves as a result of incidents that happen in space and time, and actions that are taken. The state transition probabilities are represented by the random variable $p_{ij}(a)$, which for any state $s_i$, represents the probability over the system transitioning to state $s_j$ when action $a$ is taken. Also, the time taken for the transition is represented by the random variable $t(i,j,a)$. We collectively refer to the state transition probabilities and time transition probabilities as \textit{transition probabilities} throughout the rest of the paper. 

% and point out some of the key insights about them. First, we highlight that the occurrence of decision-making states is governed by the occurrence of incidents, time taken to service incidents as well travel patterns of the responders. This makes the transition probabilities unknown and difficult to compute in closed-form. In such a scenario, one way to find the transition probabilities is to empirically estimate their values by using a simulator \cite{mukhopadhyayAAMAS18}. This process is computationally expensive even for a problem scenario that does not capture the environment explicitly as part of the state space \cite{mukhopadhyayAAMAS18}. Therefore, such an approach is clearly infeasible for our problem structure. For our model however, we bypass the need to estimate these probabilities, as we show later. We acknowledge that the state transition probabilities govern the evolution the system, but do not formally present a formulation for the same since our algorithmic framework does not need one. 

\textbf{Rewards:}
Since the broader goal of this problem is to minimize response times for emergency responders, we choose to look at \textit{costs} instead of \textit{rewards}. For each action $a$ that is taken in state $s_i$, the system incurs a cost $\rho(s_i,a)$, and we seek to find actions for each state that minimizes the expected sum of costs. 
% We point out that $\rho$ and \textit{utility} are used interchangeably through the text, for the ease of discussion.

\textbf{Policy:}
A policy for a decision-making problem specifies an action for each state of the system. The goal of solving the SMDP is to find a policy that maximizes the sum of expected rewards that the decision process generates as a result of following the said policy. Our goal is to approximate the optimal policy $\pi^{*}$ which, starting from for an arbitrary state $s_i$, minimizes the sum of expected discounted costs.

\section{Our Solution}\label{sec:solution}

 \label{sec:solutionOverview}

%\subsection{Solution Overview and Technical Challenges}

%
Before introducing the technical details of our solution approach, we provide a broad overview of the algorithmic approach we take and the associated technical challenges. We point out that two characteristics are fundamental to the operation of an emergency response system - first, it must be equipped with the ability to perform real-time computing in order to process the continuous stream of data received from responders and calls, and secondly, it must be equipped with principled algorithmic approaches to dispatch responders as and when incidents happen. We clarify that real-time computation essentially refers to a soft real-time problem - once incidents happen, the entire pipeline can afford a \textit{short} latency to update existing models and calculate dispatch decisions. 
With these characteristics in mind, we start by looking at incident prediction algorithms. The canonical way to predict incidents in space and time is using historical data to learn a predictive model and then simulate incidents \cite{mukhopadhyay2016optimal,CALIENDO2007657}. However, since accidents often cascade, it is imperative that the model is updated as and when incidents happen. The primary technical challenge here is that re-training the entire model each time an incident happens (or periodically after some pre-defined number of incidents) is computationally slow and puts a heavy toll on the responder-dispatch framework that can only afford a low latency. This calls for the need to design an online mechanism to predict incidents that can be updated as incidents happen. We introduce such a model and explain the algorithmic details involved in section \ref{sec:incidentPrediction}.

Having looked at the problem of incident prediction, we now look at dispatching responders given an incident prediction model. The SMDP formulation introduced in section \ref{Sec:SystemModel} is difficult to solve since the state transition probabilities are unknown and cannot be computed in closed-form. One way to tackle this problem is to access a generative model to learn the state-transition probabilities while learning a policy. This has recently been shown to work well on the responder-dispatch problem \cite{mukhopadhyayAAMAS18}. However, we point out that such an approach has two major limitations. First, for any urban system, the state space is practically intractable even without environment variables. Even on fairly powerful computing systems, it would take weeks to train the policy \cite{mukhopadhyayAAMAS18}. The inclusion of environment variables would be computationally infeasible. Secondly, and partly as a consequence of the first issue, prior approaches are simply not suited for dynamic environments: if a single responder breaks down, traffic conditions change, or incident models evolve, existing approaches \cite{mukhopadhyayAAMAS18,keneally2016markov} prescribe re-learning from scratch, which takes time that is incompatible with the latency constraints on the system. In order to alleviate this concern, we take the SMDP formulation and design an algorithmic approach that bypasses the need to learn the transition probabilities. This saves vital computation time and lets us design an online algorithm that is updated in real-time as the environment evolves (see Section \ref{sec:dispatchAlg}).

% \begin{figure}[t]
% \centering
% \includegraphics[width=\columnwidth]{Images/systemModelUpdated.pdf}
% \vspace*{-.1in}
% \caption{System Overview}
% \label{fig:systemOverview}
% \vspace{-0.3in}
% \end{figure}

\begin{figure}%[t]
\centering
\includegraphics[width=\columnwidth]{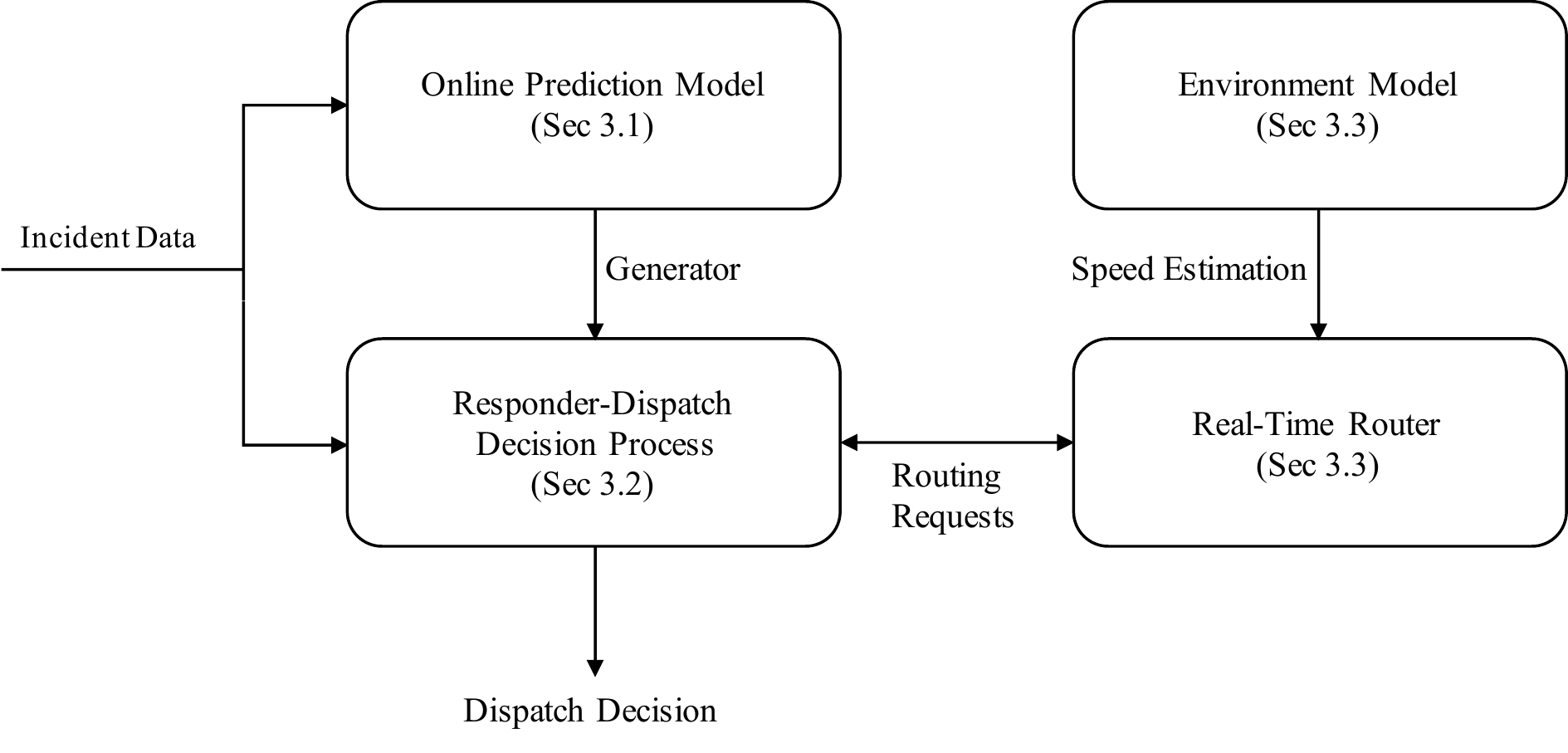}
\caption{System Overview}
\label{fig:systemOverview}
\vspace{-0.3in}
\end{figure}

%Finally, we point out that 
In order to consider the effect of environmental factors on responder dispatch, it is essential to understand how such factors evolve. This motivates us to create predictive models for the environment. One factor of interest in the context of emergency responder-dispatch is traffic conditions in urban areas, since they directly affect travel times of responders. We take this into account by describing a model that enables us to learn the evolution of traffic in an urban area from prior data, and predict traffic conditions on the fly while attempting to solve the MDP (see section \ref{sec:trafficAndRouting}).

The high-level process flow that ties the three problems into one complete pipeline of responder dispatch is shown in figure \ref{fig:systemOverview}. The online prediction model consumes actual incident data and at any point in time, provides a simulator that captures the latest trends in incident arrival. Also, the model to learn the evolution of environmental variables is used to find optimal vehicular routes between any two points in the urban area. These two atomic pieces are fed into the decision process for responder-dispatch, that given any state of the SMDP, outputs a decision that governs which responder should be sent to respond to an incident.

% This system must integrate with existing dispatching systems in use by cities. Nashville's system involves the following steps: (1) A call taker receives a 911 call, and asks questions to determine the incident type. (2) Dispatching software assigns the type, and inserts the incident into a pending queue. (3) Incident waits in queue for a dispatcher to be available. This typically takes seconds, but can be longer if dispatchers are busy. (4) Dispatcher assigns appropriate responders to the incident \cite{fireDepartmentCommunication} . Our system can start processing the incident as soon as it is put into the queue in step (2), and must have it's recommendation ready by the time a dispatcher decides on responders in step (4).

\subsection{Real Time Incident Prediction} \label{sec:incidentPrediction}

We now present the technical details and formalize our methodology, and begin by looking at a principled incident prediction algorithm.
%, since a fundamental necessity of any responder dispatch framework is to first understand where and when incidents happen.
% , and this calls for the need to design principled spatio-temporal models of incident arrival.
Formally, we want to learn a probability distribution over incident arrival in space and time. In order to do so, we leverage our prior work in which we have shown how survival models prove to be extremely effective in predicting incidents like crimes and traffic accidents \cite{mukhopadhyay2016optimal,mukhopadhyayAAMAS17,mukhopadhyayAAMAS18}. Survival Analysis is a class of methods used to analyze data comprising of time between incidents of interest \cite{cox1984analysis}. Survival models can be parametric or non-parametric in nature, with parametric models assuming that \textit{survival time} follows a known distribution. 
%Moreover, one may choose to model either the risk of occurrence of events (hazard) or the time to occurrence of incidents as the dependent variable of interest. For spatial-temporal models however, it is natural to model the time between incidents, and specifically, we use the Accelerated Failure Model (AFT) for the survival function that lets us directly model time to incidents $\tau$ as a function of features $w$. 
Based on our prior work, we choose a parametric model over incident arrival, and represent the survival model as $f(\tau|\gamma(w))$, 
%\[
%, %\, t \geq 0,
%\]
where 
$f$ is the probability distribution for a continuous random 
variable $\tau$ representing the inter-arrival time, which typically depends on
covariates $w$ via the function $\gamma$. The model parameters can be estimated by the principled procedure of Maximum Likelihood Estimation (MLE). The spatial granularity at which such models are learned can be specified exogenously - a system designer can choose to learn a separate $f$ for each discretized spatial entity (grids in our case), learn one single model for all the grids or learn the spatial granularity from data itself. This choice is orthogonal to the approach described in this paper and we refer interested readers to our prior work \cite{mukhopadhyayAAMAS17} for a discussion about such models.

We shift our focus directly to survival models that are used to learn $f(\tau|\gamma(w))$. Intuitively, given an incident and a set of features, we want to predict when the next incident might happen. Before proceeding, we introduce an added piece of notation - we assume the availability of a dataset $D$ $= \{(x_1,w_1), (x_2,w_2), ..,(x_n,w_n)\}$, where $x_i$ represents the time of occurrence of the $i^{\text{th}}$ incident and $w_i$ represents a vector of arbitrary features associated with the said incident. A realization of the random variable $\tau$, used to measure the inter-arrival time between incidents, can be represented as $\tau_i$ $=$ $x_{i+1} - x_i$. The function $\gamma$ is usually logarithmic and the relationship of the random variable $\tau$ with the covariates can be captured by a log-linear model. Formally, for a time-interval $\tau_i$ and associated feature vector $w_i$, this relationship is represented as

\begin{equation}\label{eq:logLinear}
\begin{aligned}
    &log(\tau_i) = \beta_1 w_{i1} + \beta_2 w_{i2} + ... + \beta_m w_{im} + y
\end{aligned}
\end{equation}

where, $\beta \in \mathbb{R}^{m}$ represents the regression coefficients and $y$ is the error term, distributed according to the distribution $h$. The specific distribution of $f$ is decided by how the error $y$ is modeled. We choose to model $\tau$ by an exponential distribution
% An exponential distribution for $t$ is particularly helpful in creating the SMDP model described in section \ref{Sec:SystemModel} as well as for sampling incidents 
(for the sake of brevity, we refer interested readers to prior work \cite{mukhopadhyayAAMAS18} for more details on why an exponential distribution is particularly useful in such models). It turns out that when $y$ follows the extreme value distribution, then $\tau$ is distributed exponentially. Thus, in our incident prediction model, we assume that $h$ takes the following form

\vspace{-0.1in}
\[
    h_Y(y) =e^{y - e^{y}} 
\]

Using equation \ref{eq:logLinear}, for a given set of incidents, the log-likelihood of the observed data under the specific model can be expressed as 

\begin{equation}\label{eq:logLikelihood}
\begin{aligned}
    &L = \sum_{i=1}^{n} \text{log}\,\, h(\tau_i - w^{T}_i \beta)
\end{aligned}
\end{equation}

The standard way to estimate the parameters of the model is to use a gradient-based iterative approach like the Newton-Raphson algorithm, yielding a set of coefficients $\beta^{*}$ that maximize the likelihood expression. Now, the model over inter-arrival times is generative, it can be used to simulate chains of incidents, which is particularly helpful in building a simulator, the purpose of which we explain in the next section.

% The canonical way in which such a prediction mechanism works is \textit{offline}, meaning that the model is trained on a set of historical data and is subsequently used as a predictive tool. 

As pointed out before, such an approach is \textit{offline}. However, it is imperative to capture the latest trends in incident arrival to accurately predict future incidents, which motivates us to design an online approach for learning and predicting incidents. We introduce some added notation before describing the algorithmic approach. First, we reiterate that $\beta^{*}$ is used to refer to coefficients already learned from dataset $D$.
% Also, we overload the notation for $\beta$, referring to the vector of coefficients as $\beta$ as described and the
% Also, we refer to the $i^{\text{th}}$ element of the vector $\beta$ by $\beta_i$. 
Further, we assume that a new set of incidents $D^{'}$ $= \{(x^{'}_1,w_1),(x^{'}_2,w_2),..,(x^{'}_k,w_k)\}$ is available that consists of incidents that have happened after (in time) the original set of incidents. We aim to update the regression coefficients $\beta$ using $D^{'}$, assuming that the model already has access to $\beta^{*}$. 

In order to address this problem, we use stochastic gradient descent to update the distribution $f$ in an online fashion.  Formally, we start with the known coefficients $\beta^{*}$ and, at any iteration $p$ of the process, we use the following update rule

\[
    \beta^{p+1} = \beta^{p} + \alpha \nabla L(\beta^p,D^{'})
\]

where $\nabla (L(\beta^*,D^{'})$ is the gradient of the log-likelihood function calculated using $D^{'}$ at $\beta^p$ and $\alpha$ is the standard step-size parameter for gradient based algorithms. Using equation \ref{eq:logLikelihood}, likelihood of the incidents in the dataset $D^{'}$ can be represented by

\[
 L = \sum_{i=1}^{k} \text{log}\,\, e^{(\text{log} \tau_i - \beta^{*}w) - e^{(\text{log} \tau_i - \beta^{*}w)}}\\
\]

and subsequently, 

\[
	\frac{\partial L}{\partial \beta_j} = \sum_{i=1}^{k} -w_{ij} + w_{ij}\{ e^{(\text{log} \tau_i - \beta^{*}w_i )}  \}
\]

% The update rule can therefore be written as 

% \[
%     \beta^{p+1} = \beta^{p} + \alpha \sum_{i=1}^{k} -w_{ij} + \{ e^{(\text{log} t_i - \beta^{*}w_i )}  \}
% \]

The update step is repeated until improvements in the likelihood of the available data. Having already summarized the important steps in the algorithm in this section, we present it formally in Algorithm \ref{streamingAlgo}.

% \begin{algorithm}[H]
% \caption{Streaming Survival Analysis}
% \label{streamingAlgo}
% \SetKwProg{StraemingSurv}{Function \emph{Streaming Survival Analysis}}{}{end}

% Map store=new Map(obj, queue)\;

% \generate{Object pivot}{
%      \ForAll{child $c$ in pivot}{
%      \If{ $c$'s FieldContext is not set and $c$ is fusible}{
%           generate($c$)\;
%       }
%       \ElseIf{another condition}{
%       else if statement\;
%       }
%       \Else{
%       this is else block \;
%       }
      
%      }
%      build pivot's fieldContext $fc$\;
%      EmitClassName\;
%      EmitFields($fc$)\;
%      EmitMethods($fc$)\;
% }
% \end{algorithm}
\begin{algorithm}[t]
\small
\caption{Streaming Survival Analysis}
\label{streamingAlgo}
\textbf{INPUT}: Regression Coefficients $\beta^{*}$, Dataset $D^{'}$, Tolerance $\alpha$, Likelihood Function $L$, Maximum Iterations $MAX\_ITER$ \;

\For{$p = 1..MAX\_ITER$}{
    $\beta^{p+1} = \beta^{p} + \alpha \nabla L(\beta^p,D^{'})$ \;
    \If{$L(\beta^{p+1},D^{'}) < L(\beta^{p},D^{'})$}{
        Return $\beta^{p}$\;
    }
}
Return $\beta^{p}$ 

% \vspace{-0.2in}
\end{algorithm}

This mechanism enables us to update the incident prediction model in an online manner, saving vital computation time for the responder dispatch system. Also, this implicitly betters the dispatch algorithm by generating incident chains that capture the latest trend in incident occurrence. 
%Armed with a generative incident prediction model that can be updated online, we next tackle the problem of dispatching responders as and when incidents happen, and also explain why a generative model is particularly useful for such an approach.

% The need to capture the latest information presents us with another important challenge. Remember that the state of the MDP captures information not only about incidents and responders, but also about environmental factors that affect dispatch. 

\subsection{Dispatch Algorithm}\label{sec:dispatchAlg}

% \subsubsection{Existing Approach and Technical Challenges}

We begin the discussion on our dispatch algorithm by first explaining how the SMDP problem in formulation \ref{eq:SMDP} can be solved by canonical policy iteration. A principled algorithmic approach \cite{mukhopadhyayAAMAS18} to solve the responder-dispatch SMDP is to first convert the SMDP to a discrete-time MDP $M_d$, which can be represented as

\[
    \{S, A, \bar{p}_{ij}, \rho, V_{\beta}, \beta_{\alpha} \}
\]

where $\bar{p}_{ij}(a) = \beta_\alpha^{-1} \beta_\alpha(i,a,j)p_{ij}(a)$ is the scaled probability state transition function and $\beta_\alpha$ is the updated discount factor. The transformed MDP is equivalent to the original MDP according to the total rewards criterion \cite{mukhopadhyayAAMAS18,mdpHu}, and hence it suffices to learn a policy for $M_d$.
% and refer interested readers to prior work \cite{mukhopadhyayAAMAS18,mdpHu} for details on the conversion. 
Given such a conversion, the approach to solving the MDP involves accessing a simulator to learn the state transition probabilities for $M_d$ \cite{mukhopadhyayAAMAS18}. The algorithm, \textit{SimTrans}, an acronym for \textit{Simulate and Transform}, uses canonical Policy Iteration on the transformed MDP $M_d$, with an added computation. It tracks the states and actions encountered by the simulator and gradually builds statistically confident estimates of the transition probabilities.

This process, however, is extremely slow and fails to work in dynamic environments since any change in the problem definition (the number of responders, or the position of a depot) renders the learned policy stale. In order to tackle this problem, we first highlight an important observation - one need not find an optimal action for each state as part of the solution approach since at any point in time, only one decision-making state might arise that requires an optimal action. This difference is crucial, as it lets us bypass the need to learn an optimal policy for the entire MDP. Instead, we describe a principled approach that evaluates different actions at a given state, and selects one that is sufficiently close to the optimal action. We do this using sparse sampling, which creates a sub-MDP around the neighborhood of the given state and then searches that neighborhood for an action. In order to actualize this, we use Monte-Carlo Tree Search (MCTS).

\begin{figure}[t]
\begin{center}
\vspace{-0.1in}
\centerline{\includegraphics[width=\columnwidth]{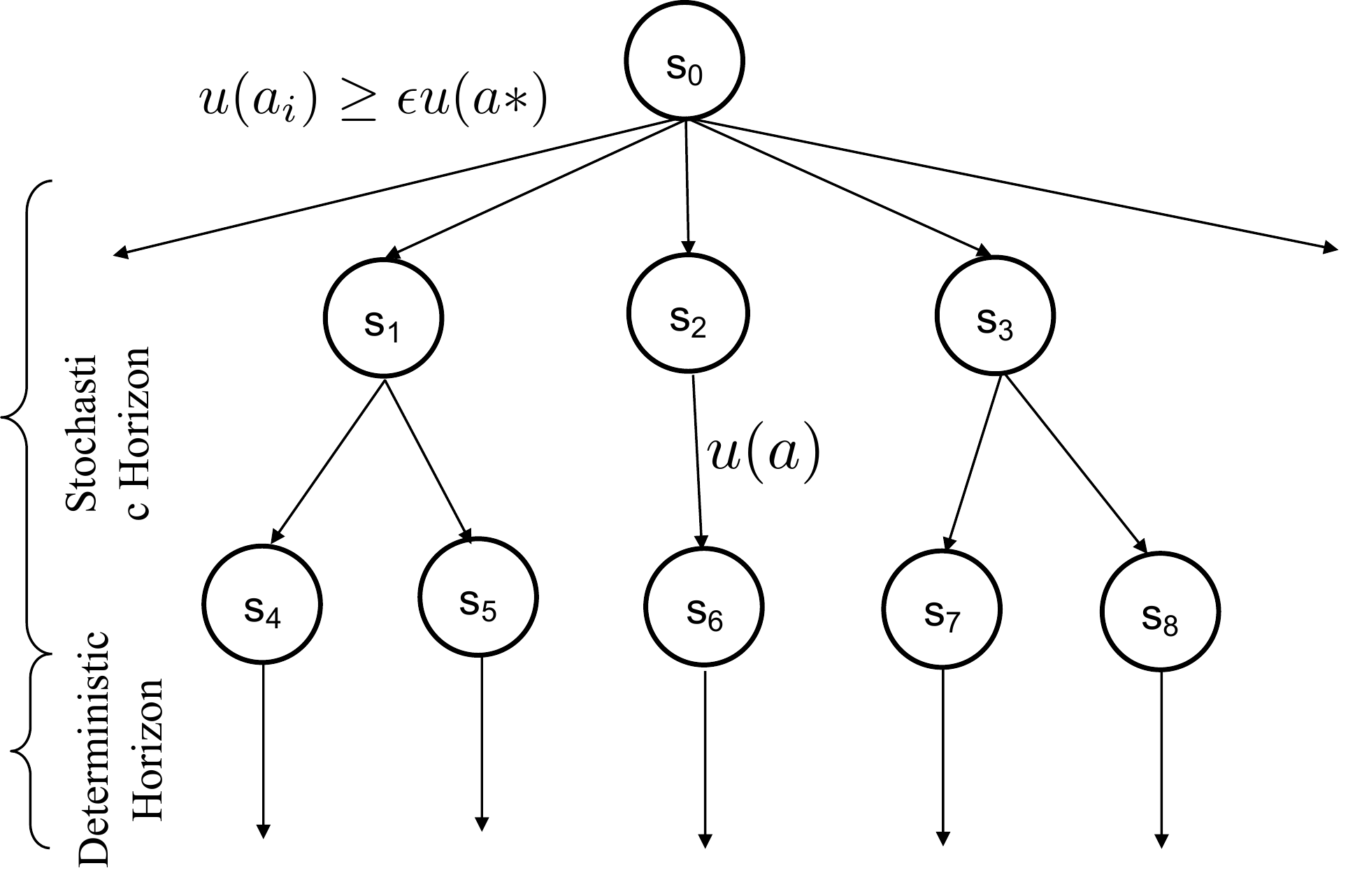}}
\caption{State-Action Tree}
\label{fig:Simdiagram}
\end{center}
\vspace{-0.3in}
\end{figure}

Another important observation is that the incident prediction model discussed in section \ref{sec:incidentPrediction} is generative and independent of dispatch decisions, which lets us simulate incidents independently.  Note that since models of travel (discussed in section \ref{sec:trafficAndRouting}) as well as service times for responders can also be learned from data, the entire urban area can therefore be simulated. We denote such a simulator by $\Theta$, which can \textit{generate} samples of how the urban area evolves, even though the exact state-transition probabilities are unknown. This observation lets us simulate future states from a given state, leading to the creation of a state-action tree as shown in Fig. \ref{fig:Simdiagram}. We use this to design an algorithmic approach called \textit{Real-Time SMDP Approximation}, and explain it next. Through the course of this discussion, we assume that the simulator can access a modular (and possibly exogenously specified) model to predict the environment at any point in time. 

\begin{algorithm}[t]
\small
\caption{Real-Time SMDP Approximation Main Procedure}
\label{alg:main_dispatch}
\textbf{INPUT}: State $s$, Current Environment $E$, Horizon $h$, Stochastic Horizon $h^s$, Simulation Budget $b$, Generative Model $\Theta$ \;

Set current depth $d \leftarrow 0$\;
$C \leftarrow b$ incident chains generated by $\Theta(E)$ \; \label{alg:main_dispatch_chains}
Set Scores $U \leftarrow \varnothing$ \;
$\bar{A}$ = \emph{SelectCandidateActions} $(s,d, h^{s})$  \; \label{alg:main_dispatch_candidate}

\ForEach{incident chain $c \in C$}{
    $u \leftarrow $ \emph{ChainEvaluation} $(c,s,d, \bar{A}, h^{s}, h)$ \;
    \ForEach{candidate action $a \in \bar{A}$}{
        $U[a] \leftarrow U[a] + u[a]$ \;
    }
}
Return $\text{argmin}_{a \in \bar{A}} (U[a])$ \;\label{alg:main_dispatch_pickAction}
%\vspace{-0.1in}
\end{algorithm}

\begin{algorithm}[t]
\caption{Select Candidate Actions for Given State}
\label{alg:candidate_actions}
\SetKwProg{SelectCandidateActions}{Function \emph{SelectCandidateActions}}{}{end}

\SelectCandidateActions{(State $s$, Depth $d$, Stochastic Horizon $h^{s}$)}{
    $A^s \leftarrow$  set of available actions in state $s$ \;
    $a^{*} = \text{argmin}_{a \in A^{s}} (\rho(s,a))$ \;
    \If{depth $d \geq h^{s}$}{
        Return $a^{*}$ \;
    } \Else{
        % Let $\bar{A}^s$ be a truncated action space s.t. \\ 
        $\bar{A}^s = \{a |  a \in A^{s}$ and $\rho(s,a) \leq \epsilon * \rho(s,a^{*})\}$ \;
        Return $\bar{A}^s$ \;
    }
}
%\vspace{-0.1in}
\end{algorithm}

\begin{algorithm}[t]
\small
\caption{Evaluate a Chain of Incidents}
\label{alg:chain_eval}
\SetKwProg{ChainEvaluation}{Function \emph{ChainEvaluation}}{}{end}

\ChainEvaluation{(Incident Chain c, State s, Depth d, Candidate Actions $\bar{A}$, Stochastic Horizon $h^{s}$, Horizon h)}{
    Set scores $u \leftarrow \varnothing$  \;
    $d \leftarrow d + 1$ \;

    \ForEach{action $a \in \bar{A}$}{
        % Calculate Next State $s^{'}$ at next incident time $c[d]$
        Next state $s' \leftarrow \emph{UpdateState}(s, a, c)$ \;
        Utility util $= s'$.responseTime \;
        Root $\leftarrow $ new Node(State = $s`$, util = util) \;
        Update $u[a] \leftarrow$ \emph{CreateStateTree}$(Root, c, d, h^{s}, h)$ \;
    }
    Return $u$
    
}
 \end{algorithm}

Algorithms \ref{alg:main_dispatch}, \ref{alg:candidate_actions}, \ref{alg:chain_eval}, and \ref{alg:state_tree} describe the various functions of our MCTS approach. We start our discussion with Algorithm \ref{alg:main_dispatch}, which is the highest level procedure that is invoked when presented with a decision-making state. First, $b$ incident chains are sampled using the generative model $\Theta$ (refer to step \ref{alg:main_dispatch_chains} in Algorithm \ref{alg:main_dispatch}), where each chain is a time ordered list of sampled incidents. We create multiple chains in order to limit the impact of variance in the generative model.  Next, the algorithm starts building the MCTS tree. We use the function $node(s,\eta,d,t)$ to refer to the creation of a node in the 
tree, which tracks the current state of the system ($s$), the cost of the path from the root to the node ($\eta$), the depth of the tree ($d$) and the total time elapsed ($t$). Also, we use \textit{UpdateState(s,a,c)} to retrieve the next state of the system, given the current state $s$, action $a$ and chain $c$. For any state, we start by finding a set of candidate actions for the given incident (refer to step \ref{alg:main_dispatch_candidate} in Algorithm \ref{alg:main_dispatch}), which takes the algorithmic flow to Algorithm \ref{alg:candidate_actions}. The candidate actions are chosen according to the current depth of the MCTS tree - if the tree is within the stochastic horizon $h^s$, the candidate actions include all actions with a cost that is at most $\epsilon$ times the cost of the myopically optimal action $a^*$. The parameter $\epsilon$ can be varied to control the trade off between the computational load of the algorithm and performance. Once the tree is deeper than $h^s$, the algorithm picks the best myopic action as a heuristic to construct the tree's nodes until depth $h$, since rewards are sufficiently discounted.
After candidate actions are found for the sampled incidents of the chain, Algorithm \ref{alg:chain_eval} is used to evaluate possible decision-making courses - each available action is tried and the MDP is simulated to generate future decision-making states, from which the entire process is repeated. This gradually builds a tree, where each edge is an action and each node is a decision-making state. We explain this procedure in Algorithm \ref{alg:state_tree}. 

The key steps of the procedure are as follows. First, costs are tracked for every branch as the tree is built (refer to steps \ref{alg:state_tree_utilCall} and \ref{alg:state_tree_utilFunction} in Algorithm \ref{alg:state_tree}), which is based on the response time in seconds for the assigned responder to the current incident. A lower cost is better, as it corresponds to lower response times. For any given node that was generated by action $a$ from parent node $p$, the cost is 

% the procedure for evaluating the candidate actions of the decision state. For each of these actions, an MCTS tree of posible future states is simulated and evaluated. The individual trees are grown by algorithm \ref{alg:state_tree}. Each candidate action is an edge in the MCTS tree, and the states that result from said actions are nodes. So, for each candidate action, a future state is simulated based on that action, including assigning the appropriate responder and updating all responder locations and status at the next decision`s time. This state is inserted in the tree as the child for the candidate action edge.  

% When a state is simulated and inserted into the tree, it is assigned an updated \textit{utility}. 

\vspace{-0.1in}
\begin{equation} \label{eq:utilUpdate}
cost = u_{p} + (\gamma ^ t)((t - u_{p})/(d + 1))
\end{equation}

where $u_{p}$ is the parent node`s cost, $\gamma$ is the discount factor for future decisions, and $t$ is the time elapsed between taking action $a$ at the parent node and the occurrence of the current node. 
% between the parent`s incident and the current incident in minutes, and $d$ is the current depth in the search tree.
This is essentially an updated weighted average of the response times to incidents given the dispatch actions. 
% In Nashville incidents happen every 57 minutes on average, so minutes are the most appropriate time unites. %\gap{support}. 

\begin{algorithm}[t]
%\vspace{-0.2in}
\small
\caption{Generate State Tree}
\label{alg:state_tree}
\SetKwProg{CreateStateTree}{Function \emph{CreateStateTree}}{}{end}
\SetKwProg{UtilityUpdate}{Function \emph{UtilityUpdate}}{}{end}

\CreateStateTree{(Parent Node n, Incident Chain c, Depth d, Stochastic Horizon $h^{s}$, Horizon $h$)}{
    \If{$d$ > horizon $h$}{
        Return n.util \;
    }\Else{
        $A = $ \emph{SelectCandidateActions}$(n.state, d, h^{s})$ \;
        $d \leftarrow d + 1$ \;
        Let ChildUtils $\leftarrow \varnothing$ \;
        
        \ForEach{candidate action $a_i \in A$}{
            Next state $s' \leftarrow \emph{UpdateState}(n.state, a, c)$ \;
            Let $\text{cost}_{i} \leftarrow \emph{UtilityUpdate}(s', n.cost, d)$ \;\label{alg:state_tree_utilCall}
            Let $x \leftarrow$ Node($s', \text{cost}_{i},d,t)$ \;
            ChildUtils $\leftarrow$ ChildUtils $\cup$ \emph{CreateStateTree}$(x, c, d, h^{s}, h)$
        }
        
        Return min (ChildUtils)
    }
        
}
\UtilityUpdate{(State s, Parent Utility $u_{p}$, Depth $d$, time $t$)}{\label{alg:state_tree_utilFunction}
% Let $\gamma$ be the Discount Factor hyper parameter \;
% Let $u_{c}$ be the response time for the current incident in $s$ \;
Return $u_{p} + (\gamma^{t})(t - u_{p})/(d + 1))$ \;
}
\end{algorithm}

% This tree generation is done recursively until depth $h^s$, which we refer to as the stochastic horizon.  
Once the tree is completed, the cost for each candidate action for the dispatch incident is determined by the cost of the best leaf node in it's sub-tree, as this represents the result of the best sequence of future actions that could be taken given the dispatch action. Finally, the algorithm averages the costs for each dispatch action across the $b$ generated incident chains, and selects the candidate action with the minimum overall cost as the best action in the current state (refer to step \ref{alg:main_dispatch_pickAction}). 

If all the responders are busy when an incident occurs, the incident is placed in a waiting queue. As soon as a responder becomes available, it is assigned to the incident at the front of the queue. This continues until the queue is emptied, after which the algorithm returns to using the heuristic policy above.

% Despite considering more responders, however, generally the response time does not decrease as $\epsilon$ increases (with the 6 responder environemnt being the only exception). This implies that when the base greedy policy is not optimal, the optimal responder is still fairly close to the incident (within $1.5$ times the closest responder). Therefore, for environments with many resonders or when using limited hardware, choosing a small value for $\epsilon$ still finds near optimal dispatching decisions while minimizing computation overhead. \\\hline

% We found $b = 10$ to be a good compromise between balancing variance and system strain, but this will greatly depend on the hardware used.

\subsection{Predicting Environmental Factors}
\label{sec:trafficAndRouting}
% \ad{put a note that though emergency vehicles have road priority we assume that their delay is still proportional to the actual travel time.}
% So far, we have looked at an incident prediction model that can seamlessly update streams of incident data in an online manner, and a responder dispatch framework that can efficiently compute an action given a state of the responder-dispatch MDP. 

We now look at the final component of the proposed pipeline - in order to capture the effect of environment, we must learn how the environment evolves. We specifically focus our attention to traffic conditions, that directly affect the movement of responders. While information about current traffic conditions can be collected while making decisions, it does not suffice for long-term planning. As the dispatch algorithm builds the state-action tree into the future, estimates of environmental variables are needed ahead of time, thereby making it imperative to learn predictive models for such variables. We therefore, design an algorithmic approach to predict future traffic conditions, and highlight how it can be used with an appropriate route-finding algorithm to predict travel times for emergency responders.

% First, we reiterate that the obvious issue in using distance between two points as a proxy for travel times is that such an approach neglects the effect of traffic conditions. In the context of emergency responders, this could be particularly disastrous as time lost in navigating through traffic congestion could eventually cause losses of life and property. 
% This makes it imperative to take traffic conditions into effect when designing responder dispatch systems, and is precisely why such information is captured by the environment variable $E$ in the MDP model.
%However, simply taking current traffic conditions into consideration does not suffice. As the dispatch algorithm builds the state-action tree into the future, such environmental variables need to be predicted to accurately capture the dynamics of the decision-process. Therefore, we describe an approach to learn and predict traffic in an urban area, and highlight how it can be used with an appropriate route-finding algorithm to predict travel times for emergency responders.

%   We therefore, design an algorithmic approach to predict future traffic conditions, which aids in finding the optimal dispatch decision. 

\textbf{Traffic Prediction Model:}
%\label{sec:traffic_prediction}
%We start with a model to learn and predict traffic conditions. 
We model the urban area as a set of road segments. For each segment, we assume that the dataset contains an associated set of features, which include  data about the number of lanes, length of a segment, vehicular speed at different times and so on.
% \ad{uncomment if an appendix is added.}
%(we refer the readers to the Appendix for a complete description of the features used).
Using this data set and features we learn a function over vehicular speed on a segment, conditional on the set of features using a Long Short-Term Memory Neural Network (LSTM) \cite{hochreiter1997long} model. The primary capability of such a framework is to model long-term dependencies and determine the optimal time lag for time series problems, which is especially desirable for traffic prediction in the transportation domain.

\textbf{Route Finding Algorithm:}
%\label{sec:router}
Armed with a model that can predict vehicular speed on road segments, we now look for an approach to find the optimal route between two given points in the urban area. Specifically, given a source, destination and departure time, we seek to find the route with minimum \emph{expected} travel time. To this end, we design a router based on  A$^{*}$ search with landmarks (\textit{ALT}) \cite{goldberg2005computing}. \textit{ALT} improves upon euclidean-based A$^{*}$ search \cite{hart1968formal} by introducing landmarks to compute feasible potential functions using the triangle inequality, thereby improving the computational cost involved with such a procedure.

\section{Performance}\label{Sec:Experiments}

\begin{table}[t]
\caption{Final Hyper-Parameter Choices} 
% \small
\vspace{-0.1in}
\footnotesize
\begin{tabular}{|p{3.5cm}|l|l|l|l|}
\hline
\begin{tabular}[l]{@{}l@{}}Number of Stations \\ (Fraction of Nashville Count) \end{tabular}%Number of Responders (Fraction of full Nashville Count)
    & \begin{tabular}[l]{@{}l@{}} 26 \\ (full) \end{tabular}%26 (full)
    & \begin{tabular}[l]{@{}l@{}} 13 \\ (1/2) \end{tabular}%13 (1/2) 
    & \begin{tabular}[l]{@{}l@{}} 6 \\ (1/4) \end{tabular}%6  (1/4)    
    & \begin{tabular}[l]{@{}l@{}} 3 \\ (1/8) \end{tabular}%3  (1/8)   
    \\
    \hline
    
Simulation Budget $b$
    & 10              % full - 5/1386
    & 10        % 13 - 14/1386
    & 10       % 6 - 99/1386
    & 10       %3  - 150/1386
    \\
    \hline
    
Candidate Action Factor $\epsilon$
    & 1.5              % full
    & 1.5        % 13
    & 2.5      % 6
    & 1.5       %3  - 
    \\
    \hline
    
Stochastic Horizon $h^{s}$
    & 1              % full
    & 1        % 13
    & 2      % 6
    & 1       %3  
    \\
    \hline
    
Discount Factor $\gamma$
    & 0.9             % full
    & 0.9        % 13
    & 0.99999       % 6
    & 0.99999       %3
    \\
    \hline
    
\end{tabular}
\label{tbl:param_choices}
\vspace{-0.2in}
\end{table}

\subsection{Data and Methodology}

Our evaluation uses traffic accident data obtained from the fire and police departments of Nashville, TN, which has a population of approximately 700,000. We trained the generative survival model on 9345 incidents occurring between 1-1-2016 and 2-1-2017, and evaluated the algorithm on 1386 incidents occurring between 2-1-2017 and 4-1-2017. We gathered information about road segments and their geographical locations using real-time traffic data collected from HERE Traffic API \cite{HEREApi} for Nashville area. The granularity of this dataset lets us access real-time vehicular speed for all segments in Nashville, which is sampled every minute throughout the day.

\textbf{Caching the Router Results:} While we recommend using a \textit{router} in real-time using the exact locations of responders and incidents to make decisions, it is not feasible for our experiments. Our preliminary analysis showed that each router request takes approximately 0.2 seconds on average.  
% Since there are many requests required to simulate responses to future incidents, even with caching routing results, it could take over 1 minute to make a response decision for an incident. This is simply too long to wait, therefore we decided to use a cached approach. 
In order to reduce the query time needed to find vehicular speed between arbitrary locations, we cached travel times between locations for different times of the day, with time discretized every 30 minutes. Our experiments showed that travel times in Nashville do not change significantly at this interval (ranging from 2-7 mph).
% Figure \ref{fig:speed_change_plot}, which plots the average change in speed each hour across Nashville`s traffic network, demonstrates that speed never deviates more than 7 mph between each hour of the day.
In order to actualize caching, we used the same grid system described in section \ref{Sec:SystemModel} for locations, with any location in the city discretized to the centroid of it's grid.

\begin{figure*}%
\centering
\subfigure[][]{%
\label{fig:ex3-a}%
\includegraphics[width=.47\textwidth]{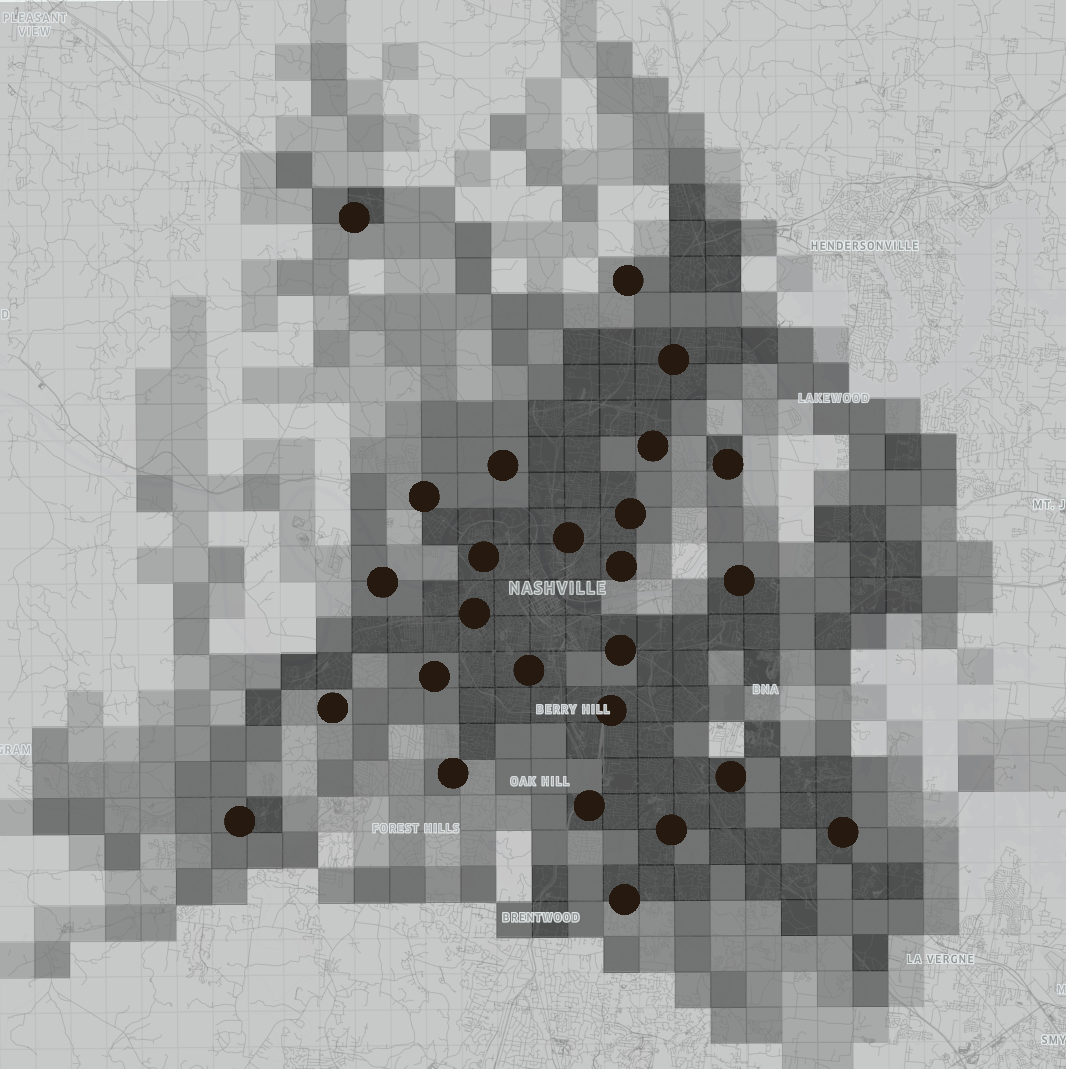}}%
\hspace{8pt}%
\subfigure[][]{%
\label{fig:ex3-b}%
\includegraphics[width=.47\textwidth]{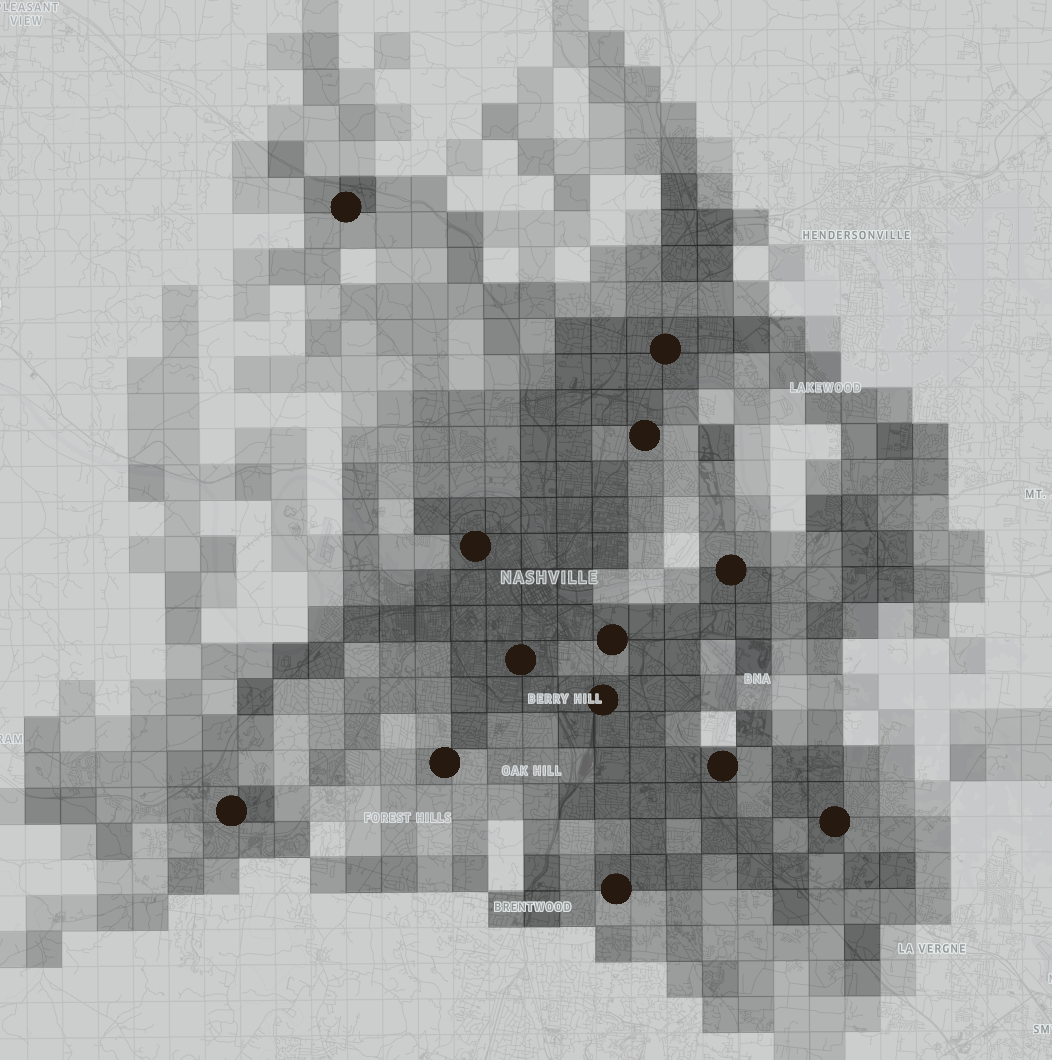}}\\
\subfigure[][]{%
\label{fig:ex3-c}%
\includegraphics[width=.47\textwidth]{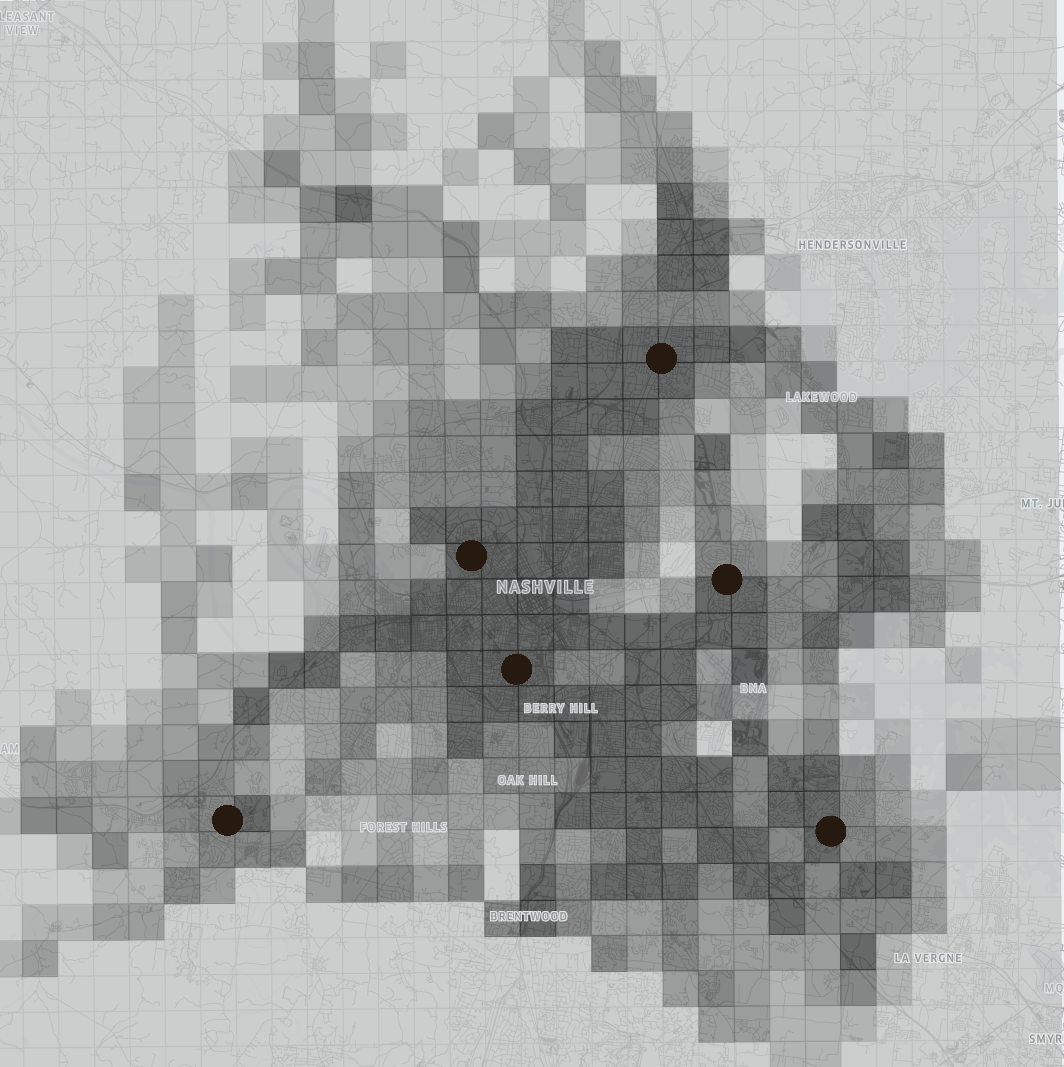}}%
\hspace{8pt}%
\subfigure[][]{%
\label{fig:ex3-d}%
\includegraphics[width=.47\textwidth]{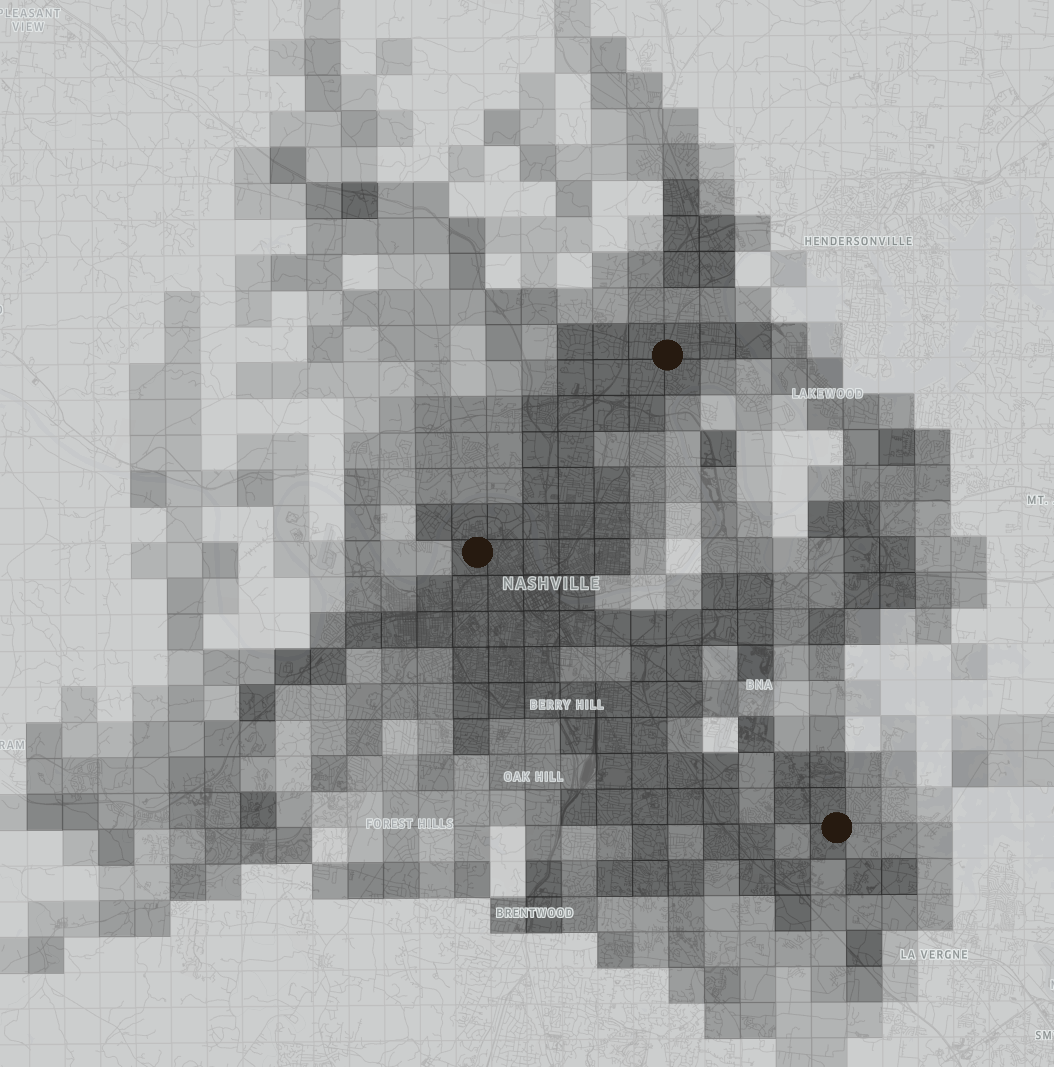}}%
\caption[A set of four subfigures.]{Distribution of the stations in each experiment overlaid on an incident occurrence heatmap (the background shows the map of Nashville, TN):
\subref{fig:ex3-a} Actual (26) stations;
\subref{fig:ex3-b} 13 stations;
\subref{fig:ex3-c} 6 stations; and,
\subref{fig:ex3-d} 3 stations.}%
\label{fig:station_heatmaps}%
\vspace{-0.2in}
\end{figure*}

\subsection{Experimental Setup}

We begin by evaluating the streaming survival model separately by comparing it to a batch-learning approach \cite{mukhopadhyayAAMAS18}. Then, we evaluate the performance of the optimizers that are used for the router, and finally evaluate the dispatch algorithm. There are two considerations that need to be made during the evaluation -

\begin{enumerate}[noitemsep,nolistsep,leftmargin=*]
    \item \textbf{Decreased Responder Availability}: It is reasonable to assume that the base policy of dispatching the closest responder is correct most of the time and it is only rarely that non-greedy actions are needed. We hypothesize that such situations occur more frequently in practice as the strain on the system is greater: i.e. the incident to responder ratio is higher. This happens since responders attend not only to traffic accidents, but to a variety of other incidents (crimes for example). To take this into account, we ran several experiments with different number of responders: The full Nashville responder count of 26, and then cutting it by a factor of half three times to simulate test-beds with 13, 6, and 3 responders. The locations of the stations in each of these test-beds compared to Nashville's incident density heatmap is shown in figure \ref{fig:station_heatmaps}.
    \item \textbf{Hyper-Parameters}: We performed a hyper-parameter search (refer to the appendix for a concise summary of the hyper-parameters) for each of the test-beds based on the number of stations. The parameters that gave the best response time savings were chosen for each set, shown in table \ref{tbl:param_choices}. We note that each hyper-parameter is important and strongly prescribe that each should be tested and tuned carefully for a new environment and hardware the system is deployed on, as these values may not be optimal for more constrained hardware, different responder distributions, or different cities with other incident arrival models.

\end{enumerate}

\vspace{0.1in}
\subsection{Results and Discussion} \label{sec:performance_results_discussion}

\subsubsection{Streaming Survival Analysis}

\begin{figure}[t]
\begin{center}
\centerline{\includegraphics[width=0.9\columnwidth]{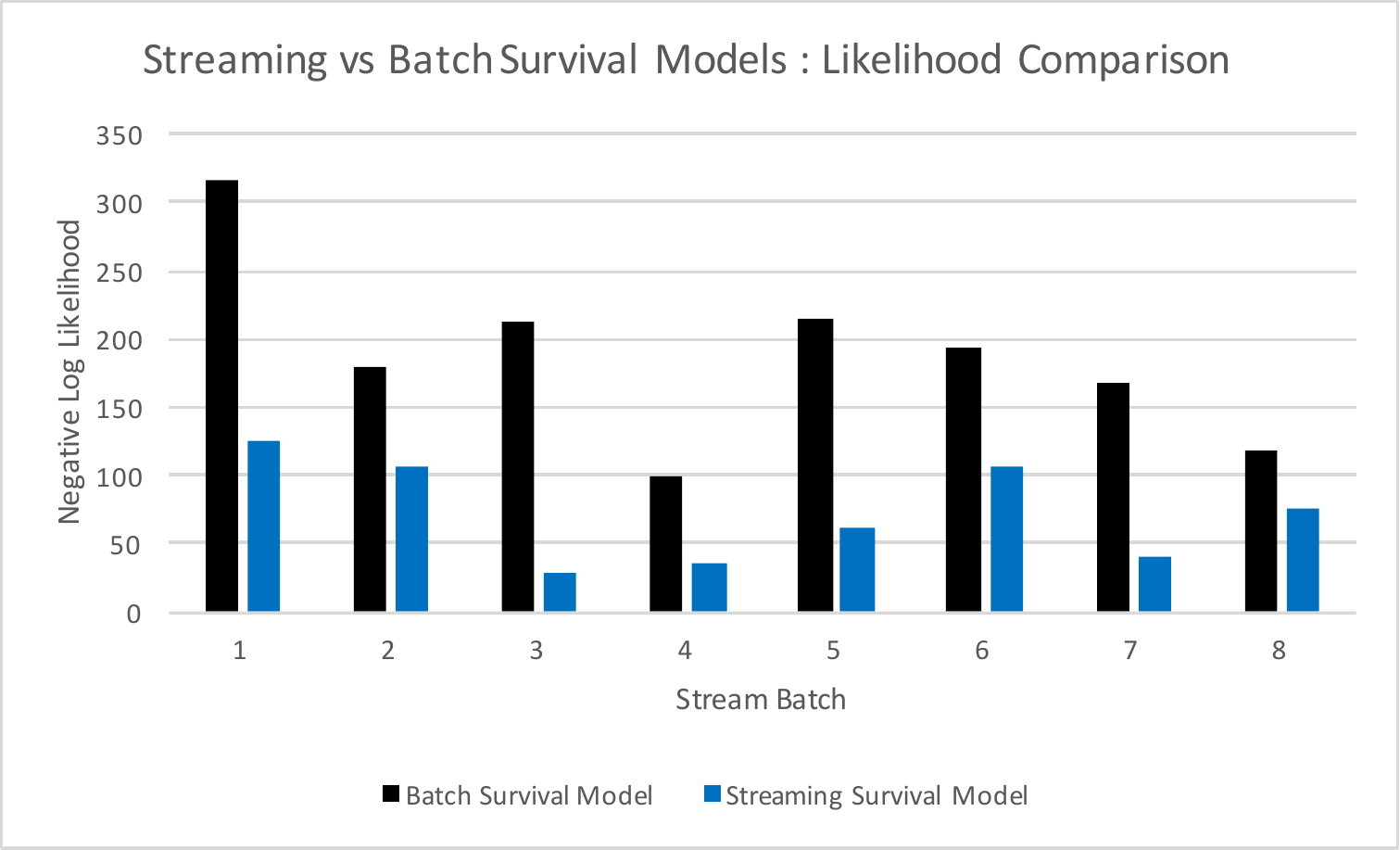}}
\caption{Negative Log Likelihood comparison (lower is better) between Batch and Streaming Survival Models. The Streaming model outperforms the batch model by a significant margin}
\label{fig:likelihoodResults}
\end{center}
\vspace{-0.3in}
\end{figure}

\begin{figure}[t]
\centering
\includegraphics[width=0.9\columnwidth]{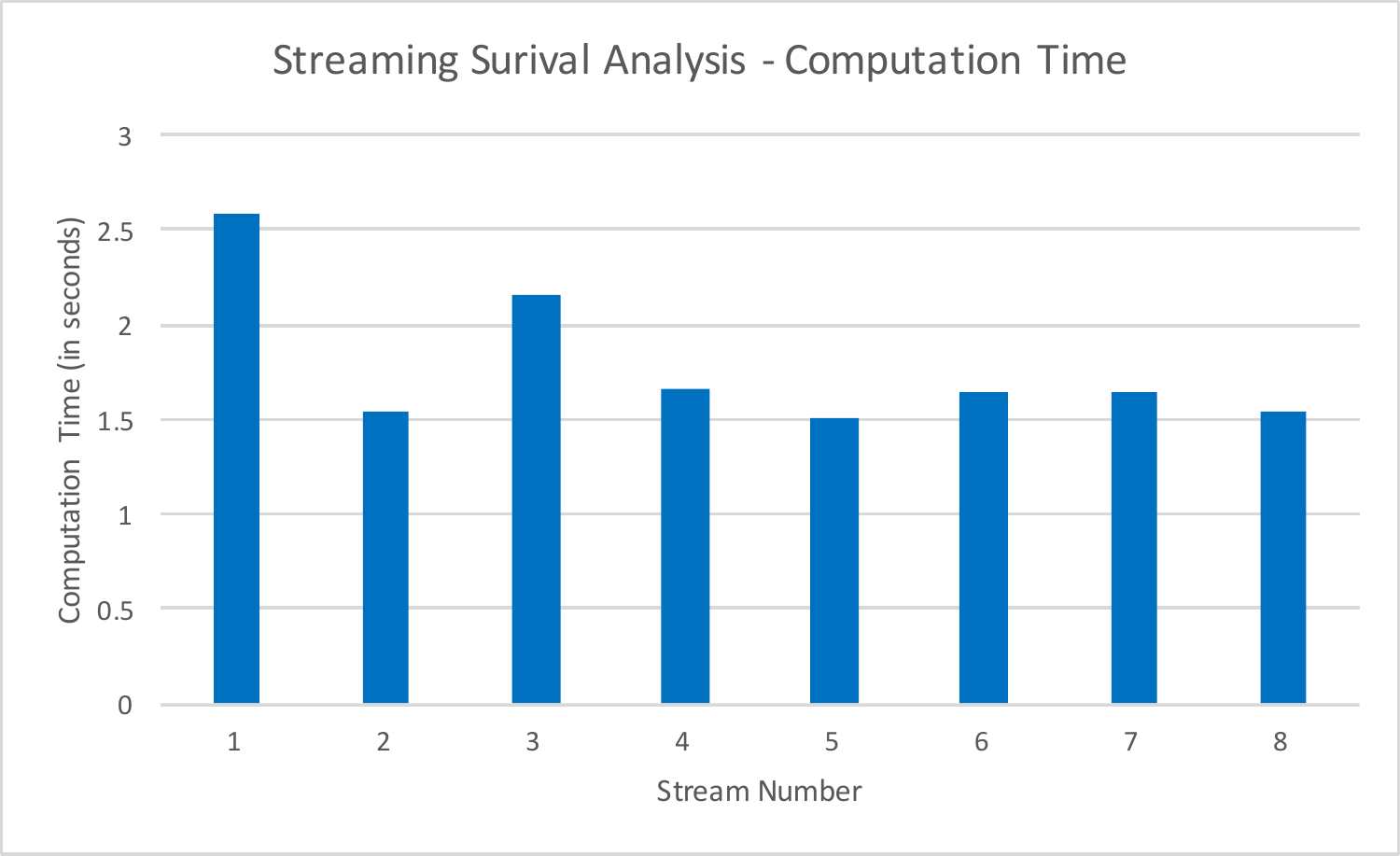}
\caption{Computational Time for Streaming Survival Analysis.}
\label{fig:runTimeStream}
\vspace{-0.2in}
\end{figure}

We learned the batch model using the entire training data set and then, in the evaluation set, we considered each week as a \textit{stream}, and further split $80\%$ of the stream into a training set and $20\%$ as the test set. We evaluated the batch model as well as the streaming model on the test set of each of the streams. Note that the batch model has access to all the data in the streams in the form of features; the stream model, on the other hand, gets updated after each data stream is received according to Algorithm \ref{streamingAlgo}. We use the \textit{de-facto} standard of comparing likelihoods for evaluation, and present the results in in Figure \ref{fig:likelihoodResults}. The streaming model results in a significant increase in likelihood (we plot the negative log likelihoods, hence lower is better) and convincingly outperforms the batch model. We point out a minor caveat - the updates can be used in practice only if the time taken to update the model is small as compared to the latency that emergency responders can afford. To illustrate this, we present the computational run-times of the stream model (for each stream) in Figure \ref{fig:runTimeStream}, and observe that it usually takes less than 2 seconds for an update to run, which justifies the usage of such models in practice.

\begin{figure*}[]
\begin{center}
\centering
\centerline{\includegraphics[width=.95\textwidth]{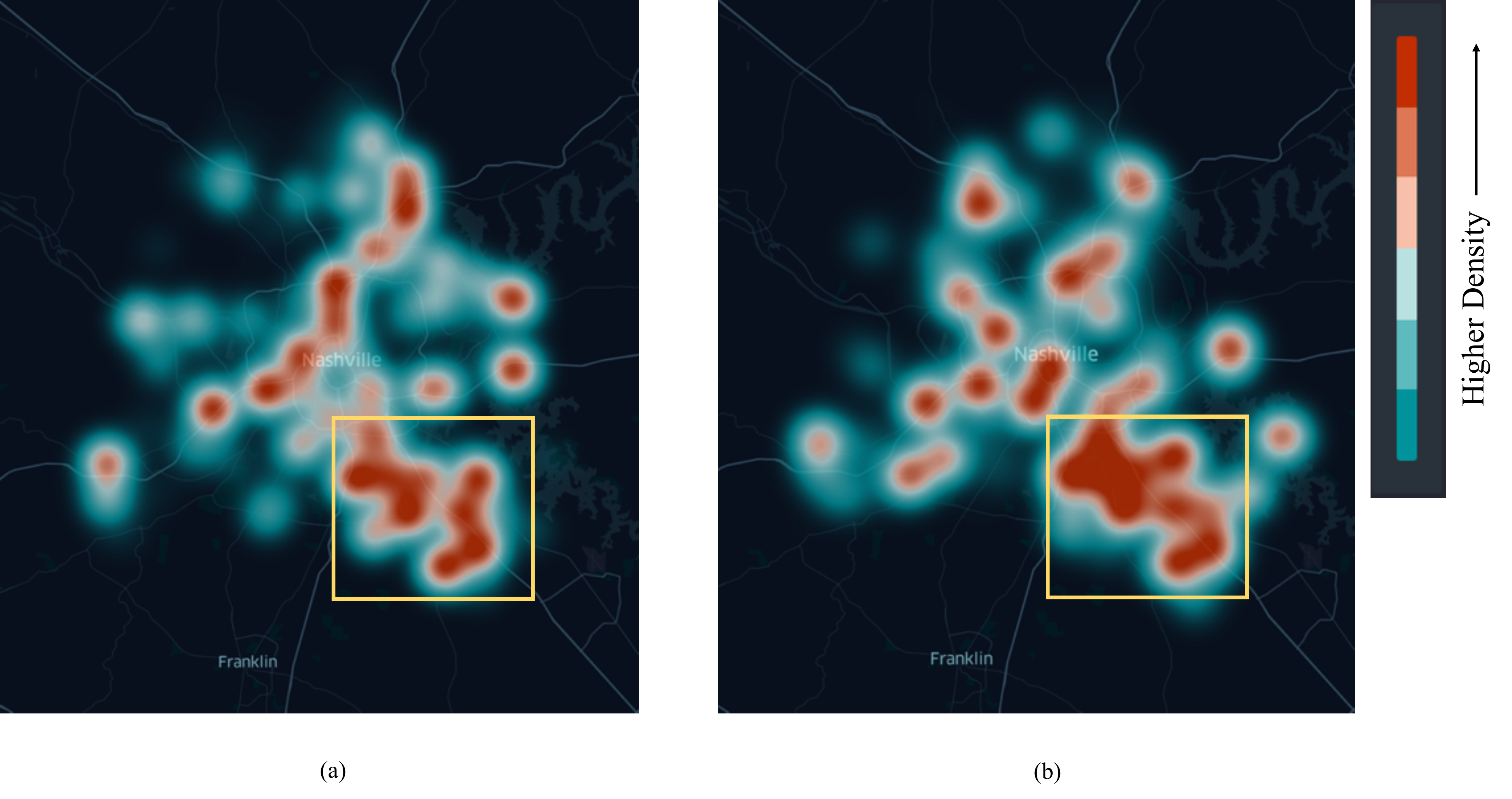}}
\caption{Batch Model (a) vs Streaming Model (b): These predicted heatmaps demonstrate that the streaming model adjusts more quickly to new incident distributions. Starting with a survival model learned from the training set, we fed the models synthetic incident data with incidents only occurring in the yellow boxed area, which means that the model should learn that there is now a higher incident likelihood in this area. The batch model picks up the new pattern weakly, whereas the streaming model shows higher likelihood in marked box.}
\label{fig:toySurvival}
\vspace{-0.3in}
\end{center}
\end{figure*}

In order to visually illustrate the benefit of a streaming model, we look at a fabricated example, where we feed the incident prediction models with data that is deviant from standard accident patterns. We show these results as heatmaps in Figure \ref{fig:toySurvival}. Note that a brighter color corresponds to higher density of incidents. We see that the batch model \textit{weakly} learns the pattern since it has access to the updated dataset only in the form of features; the streaming model on the other hand, identifies the current trend and predicts higher density of incidents in the concerned region, thereby highlighting the importance of such models in dynamic environments.

% \begin{figure*}[t]
% \begin{center}
% \centering
% \centerline{\includegraphics[height=4cm]{Images/streamingToy.pdf}}
% \caption{Streaming Model vs Batch Model : when fed with fabricated incident data (a), the batch model picks up the pattern weakly (b), whereas the streaming model shows higher likelihood in marked box (c). Brighter colors indicate higher density}
% \label{fig:toySurvival}
% \vspace{-0.1in}
% \end{center}
% \end{figure*}

\subsubsection{Predicting Travel Times}

 We briefly show results of our traffic router before moving to the dispatch algorithm. We compared the LSTM using three different optimizers (Adam \cite{kingma2014adam}, SGD \cite{robbins1985stochastic}, and Adagrad \cite{duchi2011adaptive}), and model performance was evaluated using five-fold shuffled cross validation. Adam, SGD and Adagrad showed Mean Absolute Errors of 5.47, 4.27 and 6.16 miles/hours respectively.
%  Table \ref{table:lstm_tuning_table} shows Mean absolute error (MAE) in miles/hour units for different optimizers. 
Therefore, we chose SGD for our router described in Section \ref{sec:trafficAndRouting}. While evaluating the router on unseen data, the model with SGD optimizer showed MAE of only \textbf{6.419 miles/hour}.
 
% \begin{table}[]
% \begin{tabular}{|c|c|}
% \hline
% \textbf{Optimizer} & \multicolumn{1}{l|}{\textbf{MAE}} \\ \hline
% Adam               & 5.47                              \\ \hline
% SGD                & 4.27                              \\ \hline
% Adagrad            & 6.16                              \\ \hline
% \end{tabular}
% \end{table}

% \begin{figure*}[t]
% \centering
% \centerline{\includegraphics[width=0.4\columnwidth]{Images/RunTimeSurvival.pdf}}\label{fig:runTimeStream}
% \caption{Negative Log Likelihood comparison (lower is better) between Batch and Streaming Survival Models. The Streaming model outperforms the batch model by a significant margin}
% % \vspace{-0.3in}
% \end{figure*}

\begin{figure}[t]
    \centering
\begin{tikzpicture}
\begin{axis}
[
      width=.7\columnwidth,height=4.5cm,
    %   height=0.23\textwidth,
    %   font=\footnotesize,
    %   grid=major,
    %   xtick=,
      ylabel={Number of Stations},
      title={Positively Impacted Incidents},
      ytick={1, 2, 3, 4},
      yticklabels={{3}, {6}, {13}, {26 (all)}},
      xlabel style={text width=0.8\columnwidth, align=center},
      xlabel={Response Time Decreases (Minutes)},
      ylabel near ticks
    ]
      \addplot+[boxplot={draw position=1}] table [col sep=comma, y expr=\thisrow{time}*(-1/60)] {data/3_stations_better.csv};
      \addplot+[boxplot={draw position=2}] table [col sep=comma, y expr=\thisrow{time}*(-1/60)] {data/6_stations_better.csv};
      \addplot+[boxplot={draw position=3}] table [col sep=comma, y expr=\thisrow{time}*(-1/60)] {data/13_stations_better.csv};
      \addplot+[boxplot={draw position=4}] table [col sep=comma, y expr=\thisrow{time}*(-1/60)] {data/26_stations_better.csv};
\end{axis}
\end{tikzpicture}
\begin{tikzpicture}
\begin{axis}[
      width=.7\columnwidth,height=4.5cm,
      ylabel={Number of Stations},
      ytick={1, 2, 3, 4},
      title={Negatively Impacted Incidents},
      yticklabels={{3}, {6}, {13}, {26 (all)}},
      xlabel style={text width=0.8\columnwidth, align=center},
      xlabel={Response Time Increases (Minutes)}
      ]
      \addplot+[boxplot={draw position=1}] table [col sep=comma, y expr=\thisrow{time}*(1/60)] {data/3_stations_worse.csv};
      \addplot+[boxplot={draw position=2}] table [col sep=comma, y expr=\thisrow{time}*(1/60)] {data/6_stations_worse.csv};
      \addplot+[boxplot={draw position=3}] table [col sep=comma, y expr=\thisrow{time}*(1/60)] {data/13_stations_worse.csv};
      \addplot+[boxplot={draw position=4}] table [col sep=comma, y expr=\thisrow{time}*(1/60)] {data/26_stations_worse.csv};
\end{axis}
\end{tikzpicture}
\caption{Incident response time difference between the base policy and our solution. The left chart shows the distribution of response time decreases for positively impacted incidents, while the right chart shows increases for negatively impacted incidents. The distribution of time difference in minutes (x axis) is compared across each experiment involving the various station counts (y axis).
%\textcolor{red}{fill caption}
% Results from Experiment 2.   The horizontal axis is the elapsed time in seconds. The values shown from bottom to top are: (a) time to register job offer, (b) time to register resource offer, (c) time to match offers (measured from when offers were issued), (d) time to register all actors, (e) time to select mediators before posting job offer, (f) time to complete job (measured from when the job was first posted), and (g) the execution time of the job.
}
\label{fig:resp_comp}
\vspace{-0.2in}
\end{figure}
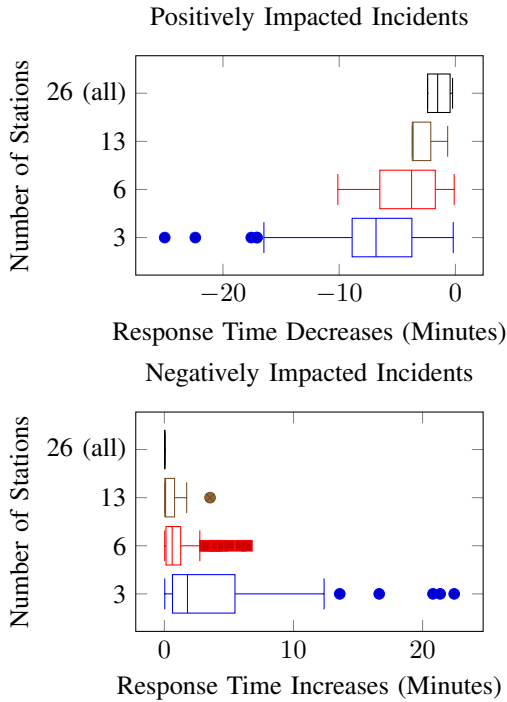

\subsubsection{Responder Dispatch} In table \ref{tbl:comparison_performance} we present the results of comparing the tuned algorithms for each stations configuration. We compare our solution against the base policy (sending the nearest responder) using the average response time reduction for incidents impacted by the algorithm (i.e. incidents with different response times than the base policy), the number of incidents impacted, and the average computation time. The first observation is that the computation times are all well within acceptable limits, as they are near the human decision maker's visual reaction times \cite{taokaReaction1989}. This demonstrates that the system overcomes the technical challenge of running and updating in real-time, and can integrate into emergency response systems described in section \ref{sec:intro}.

We observe that when there is high responder coverage, demonstrated by the experiment with 26 stations, the baseline policy is nearly always used, with only 5 of the 1386 incidents serviced being impacted by the policy. But as the number of responders decreases, the baseline policy is sub-optimal for an increasing number of incidents, capping at 150 with 3 stations. This shows that the system can respond to changing responder availability, and that it is most useful when the system is strained by high incident demand. 

Last, the average time saved for impacted incidents is significant, particularly for the experiments with 26 and 3 stations, as 30 seconds can be the difference between mortality and survival in response situations \cite{blackwell2002response}. However, these represent average savings, and to dissect the performance of our approach, we plot the distributions of the response time savings for incidents that benefited from our solution, and response time increases for negatively impacted incidents in figure \ref{fig:resp_comp}.
% an average - for some incidents you must sacrifice response time to gain time responding to other incidents, otherwise it would be the greedy policy. We examine this trade-off in more

\begin{figure}
    \centering
\begin{tikzpicture}
\begin{axis}
[
    width=.7\columnwidth,height=5cm,
    title={Base Policy},
    ylabel={Number of Stations},
      ytick={1, 2, 3, 4},
      yticklabels={{3}, {6}, {13}, {26 (all)}},
      xlabel style={text width=0.4\textwidth, align=center},
      xlabel={Response times (Minutes)},
      ylabel near ticks
    ]
      \addplot+[boxplot={draw position=1}] table [col sep=comma, y expr=\thisrow{time}*(1/60)] {data/resp_time_data/Send_Closest_Resp_Times_3.csv};
      \addplot+[boxplot={draw position=2}] table [col sep=comma, y expr=\thisrow{time}*(1/60)] {data/resp_time_data/Send_Closest_Resp_Times_6.csv};
      \addplot+[boxplot={draw position=3}] table [col sep=comma, y expr=\thisrow{time}*(1/60)] {data/resp_time_data/Send_Closest_Resp_Times_13.csv};
      \addplot+[boxplot={draw position=4}] table [col sep=comma, y expr=\thisrow{time}*(1/60)] {data/resp_time_data/Send_Closest_Resp_Times_26.csv};
      
\end{axis}
\end{tikzpicture}
\begin{tikzpicture}
\begin{axis}
[
    width=.7\columnwidth,height=5cm,
    title={Our Solution},
    ylabel={Number of Stations},
      ytick={1, 2, 3, 4},
      yticklabels={{3}, {6}, {13}, {26 (all)}},
      xlabel style={text width=0.4\textwidth, align=center},
      xlabel={Response times (Minutes)}
    ]
      \addplot+[boxplot={draw position=1}] table [col sep=comma, y expr=\thisrow{time}*(1/60)] {data/resp_time_data/Best_Policy_Resp_Times_3.csv};
      \addplot+[boxplot={draw position=2}] table [col sep=comma, y expr=\thisrow{time}*(1/60)] {data/resp_time_data/Best_Policy_Resp_Times_6.csv};
      \addplot+[boxplot={draw position=3}] table [col sep=comma, y expr=\thisrow{time}*(1/60)] {data/resp_time_data/Best_Policy_Resp_Times_13.csv};
      \addplot+[boxplot={draw position=4}] table [col sep=comma, y expr=\thisrow{time}*(1/60)] {data/resp_time_data/Best_Policy_Resp_Times_26.csv};
      
\end{axis}
\end{tikzpicture}

\caption{Response time distributions of the base policy compared to our solution for each station experiment. The two are very similar on average since most incidents have the same responder, demonstrating that our solution is not worse than the base policy. The benefits of our solution are clear when looking at the response times for incidents with different dispatching decisions, as shown in figure \ref{fig:resp_comp}.}
\label{fig:raw_resp}
\vspace{-0.1in}
\end{figure}
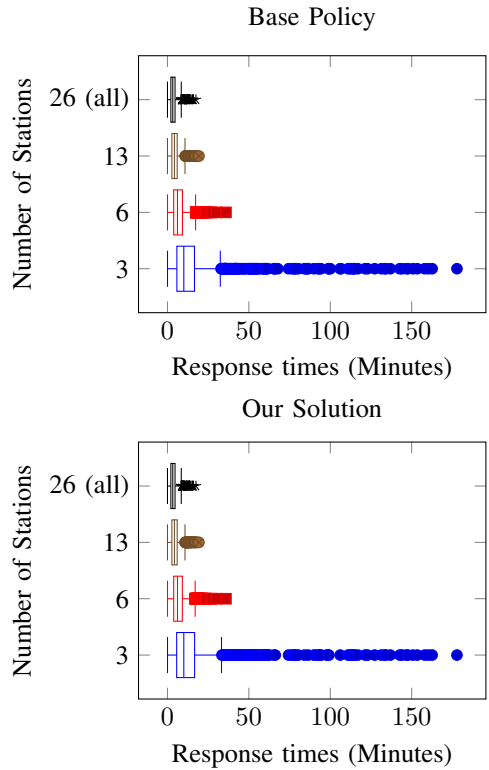

Comparing the box plots, the negatively impacted response times are more dense near zero compared to the savings. This shows that in general, the algorithm is not making large sacrifices for individual incidents in comparison to the savings generated, which is reinforced by the overall distribution of response times shown in figure \ref{fig:raw_resp}. The response times for the positively impacted incidents are generally much improved; the median improvement is over 200 seconds for the experiment with 13 stations, for example.
Unfortunately, however, there are some outliers with unacceptably large sacrifices. For example, there is an incident in the experiment with 13 stations that took over 200 additional seconds to respond to compared to the base policy, which significantly increases the potential mortality of that incident if it is severe. This raises the question of integrating severity of incidents into the SMDP model, and we plan to consider the integration of prioritization of incidents in future work. 

%Note that  

%If the severity of a given incident is known, it can determine it`s priority in the dispatching decision. Incidents with low severity and mortality risk (e.g. a minor traffic collision on a residential road) might be able to withstand increased response times without large penalties, allowing for the system to prioritize predicted future incidents with higher severity where response time is more critical (e.g. a high speed highway collision). 

%In section \ref{sec:futureWork} we discuss a potential way to address this problem by evaluating the severity of incidents, which we do not address in this paper. 

%%%%%%%%%%%%%%%%%%%%%%%%%%%%%%%%%%%%%%%%%%%%%%%%%%%%%%%%%%%%%%%%%%%%%%%
\begin{table}
\caption{Performance of System Compared to Base Policy}
\footnotesize
\begin{tabular}{|p{3.5cm}|l|l|l|l|}
\hline

\begin{tabular}[l]{@{}l@{}}Number of Stations \\ (Fraction of Nashville Count) \end{tabular}%Number of Responders (Fraction of full Nashville Count)
    & \begin{tabular}[l]{@{}l@{}} 26 \\ (full) \end{tabular}%26 (full)
    & \begin{tabular}[l]{@{}l@{}} 13 \\ (1/2) \end{tabular}%13 (1/2) 
    & \begin{tabular}[l]{@{}l@{}} 6 \\ (1/4) \end{tabular}%6  (1/4)    
    & \begin{tabular}[l]{@{}l@{}} 3 \\ (1/8) \end{tabular}%3  (1/8)   
    \\
    \hline
    % \hline

Average Response Time Savings for Incidents Impacted by Policy (seconds)
    & 38.705              % full
    & 2.231        % 13
    & 15.917       % 6
    & 34.871       %3
    \\
    \hline

Number of Incidents Impacted by Policy
    & 5              % full - 5/1386
    & 14        % 13 - 14/1386
    & 99       % 6 - 99/1386
    & 150       %3  - 150/1386
    \\
    \hline
Average Computation Time per Incident (seconds)   
    & 0.384              % full
    & 0.198        % 13
    & 0.350      % 6
    & 0.0343       %3  - 
    \\
    \hline

\end{tabular}

\label{tbl:comparison_performance}
\vspace{-0.2in}
\end{table}

\vspace{-0.1in}
\section{Related Work}\label{Sec:PriorWork}

% As pointed out, the problem of optimally allocating responders in anticipation of incidents consists of several sub-problems, which have mostly been studied in an orthogonal manner.
% \ad{there are missing citations in this section}
A Traffic Incident Management Decision Support System is an information system that supports the process of preparing for, responding to, and managing the effects of traffic incidents. It must support functions such as stationing emergency response resources, dispatching to incidents in real time, and routing resources \cite{zografos2002real}. Most of these sub-problems have been studied in an orthogonal manner. We look at prior work for each of the sub-problems. First, we look at the problem of dispatching responders given a model of incident arrival. 
% At the onset, we point out that although there is an implicit uncertainty in the system in which such responders operate, there is also an expectation of timely response, due to the critical nature of the incidents. Hence, this problem has been noted to be inherently tricky in the literature \cite{felder2002spatial}. 
Traditionally, problem has been looked at as a part of the responder allocation problem \cite{li2011covering,mukhopadhyayAAMAS17}, in which an allocation of responders to depots naturally creates an algorithm for dispatch. The problem has also been studied as part of a joint optimization problem that balances distribution of resources and response times \cite{toro2013joint}. Finally, principled decision-theoretic models have also been used to study the problem \cite{mukhopadhyayAAMAS18,keneally2016markov}, that look at learning a policy of actions for all states the urban area can be in. 

% However, there are many limitations in these approaches that prevent actual deployment in practice. Most of the approaches consider the environment in which such responders operate as static; the environment however, is dynamic and the assumption of a static environment fails to capture the dynamics of the actual problem. Further, there are strong distributional assumptions that do not apply in the real world \cite{keneally2016markov}, and therefore, such approaches fail to capture the true dynamics of the world in which such an algorithmic framework operates. Also, two common shortcomings of all the approaches mentioned above are the lack of a framework that can do online computation and the inability to capture real-time information about arbitrary environmental factors that affect dispatch (like traffic). Specifically, any problem definition that seeks to capture the rather complicated dynamics of urban areas, emergency responders and incidents results in an intractable state space. This makes offline approaches infeasible, as a slight change in the problem definition requires re-computing the policy, a process that even on a fairly modern computing system could take days \cite{mukhopadhyayAAMAS18}. We address these concerns systematically in this paper, by designing an approach that computes the action for a given state in an online manner. This not only saves vital computational time but also makes the system robust to changes in problem definition and allows for capturing current environmental factors that affect dispatch. 

The second sub-problem is that of predicting incidents like traffic accidents, crimes, fires and others, that need emergency response. The availability of such a mechanism is crucial to the first sub-problem as decision-theoretic approaches can be aided by mechanisms that can simulate the world in which such responders operate. This specific problem has been widely studied in the past. One of the most widely studied types of incidents is crime, and a variety of approaches \cite{kennedy2011risk,Short08} have been taken to tackle this problem. The problem of predicting traffic accidents has also received significant attention \cite{songchitruksa2006assessing,ackaah2011crash}. Recently, freeway accidents have been predicted using panel data analysis approach that predicts incidents based on both time-varying and site-specific factors \cite{qi2007freeway}. A survey of the literature on crash prediction models is presented in \cite{kiattikomol2005freeway}, which highlights the prevalence of Poisson distribution based models, and multiple linear regression approaches. There are also approaches that use clustering techniques to differentiate between incident types \cite{pettet2017incident}. Finally, there are generic approaches that can work with multiple incident types \cite{mukhopadhyayAAMAS17,mukhopadhyayGameSec16}.

\section{Conclusion}\label{Sec:Conclusion}
We designed a complete pipeline for the responder dispatch problem. We created an online incident prediction model that can consume streaming data and efficiently update existing incident arrival models. Then, we designed a framework for finding near-optimal decisions of an SMDP by using Monte-Carlo Tree Search, that bridges an important gap in literature by making such models computationally tractable. To aid the decision-making algorithm, we designed a Recurrent Neural Network architecture to learn and predict traffic conditions in urban areas. Our experiments showed significant improvements over prior work and existing strategies in both incident prediction and responder dispatch. We would like to highlight that while we treated incidents with equal severity, an interesting direction of future work involves designing the SMDP reward structure based on priorities, and directing responders based on incident prediction models that take severity into effect. 

\section*{Acknowledgement}
This work is sponsored by The National
Science Foundation under award numbers CNS1640624 and IIS1814958.  We thank our partners from Metro Nashville Fire Department and Metro Nashville Information Technology Services in this work.

% \clearpage
% \thispagestyle{empty}

\appendix
%\section{Appendix}

% \subsection{Monte Carlo Tree Search}

% Monte Carlo Tree Search (MCTS) is a class of heuristic-based search algorithms to compute the most promising actions for a given state of a decision-making problem. Given an arbitrary state of the problem, the essential focus of the MCTS is to repeatedly simulate the system under consideration to build a search tree that keeps track of the actions taken and the rewards received. The search tree is built in a manner that balances exploration (trying new or sub-optimal actions) and exploitation (greedily exploring the best action). The simulations are then used to make a decision for the root state.
%\newpage
%\newpage
\section{Appendix}

In order to ensure brevity in the main body of the paper, we move some areas of discussion to the appendix. We extend the discussion on all three components of the responder dispatch pipeline here, and start with our algorithm for predicting incidents.

\subsection{Data Sources}
\begin{table}[h]
\caption{Data Sources}
\footnotesize
\begin{tabular}{|l|p{0.15\columnwidth}|p{0.2\columnwidth}|p{0.2\columnwidth}|p{0.1\columnwidth}|}
\hline
\textbf{Type} & \textbf{Source}           & \textbf{Format}         & \textbf{Frequency} & \textbf{Range}  \\ \hline
Traffic       & HERE                      & Traffic Message Channel & One Minute       & 10/16 - Present \\ \hline
Accident      & Nashville Fire Department & JSON                    & Manually           & 02/14 - 06/17   \\ \hline
Weather       & Dark Sky                  & JSON                    &  Five Minutes    & 03/16 - Present \\ \hline
\end{tabular}
\label{tab:sources}
\end{table}

We collect static and real-time data from multiple data sources in the city of Nashville, TN.
Table \ref{tab:sources} shows the different data used in this work.

\subsection{Real-Time Incident Prediction}

While most of our approach towards learning a probability distribution over inter-arrival time between incidents is described in the main paper, we describe the features used to learn the survival model here. Our primary choice of features is governed by prior work and expert opinions, and we list the features used in our model in Table \ref{table:incidentFeatures}.

 \begin{table}[h]
 \small
 \renewcommand{\arraystretch}{1.5}
 \centering
 \caption{Features used in the incident prediction model.}
 \vspace{0.05in}
 \label{table:incidentFeatures}
 \begin{tabular}{>{\raggedright\arraybackslash}p{2.1cm} p{4.0cm}}
 \toprule
 \textbf{Feature} & \textbf{Description}\\ 
 \midrule
 Time of day & Each day was divided into 6 equal time zones with binary features for each.  \\
 Weekend & Binary features to consider whether crime took place on a weekend or not.  \\
 Season & Binary features for winter, spring, summer and fall seasons.\\
 Mean Temperature & Mean Temperature in a day\\
Rainfall & Rainfall in a day \\
Past Incidents   & Separate variables considered for each discrete crime grid representing the number of incidents in the last two days, past week and past month. We also looked at same incident measures for neighbors of a grid. \\
 \bottomrule
 \end{tabular}
 \end{table}

\subsection{Dispatch Algorithm}

The original problem $M_s$, as formulated in section \ref{sec:dispatchAlg} is a Markov-Decision Process which can be defined as 

\[
    \{S,A,p_{ij}(a),t(i,j,a),\rho(i,a) ,\alpha\}
\]

For any state $s_i$ and policy $\pi$, we define expected discounted total reward over an infinite horizon as 

\begin{equation}\label{eq:expectedRewards}
\begin{aligned}
	& V^{\pi} (s_i) = \sum_{n=0}^{\infty} \mathbb{E} \{e^{-\alpha T_n} \rho(s^n,\pi(s^n)) \}
\end{aligned}
\end{equation}

where $s^n$ is the state at $n^{\text{th}}$ decision epoch, and $T_n$ its duration. The broad goal of solving a general MDP is to learn a policy that maximizes the sum of expected rewards for any given state. We look at costs instead of rewards, and seek to find actions for a given state that minimizes the sum of expected costs.

The evolution of this system can be described by the following four steps:
\begin{enumerate}
\item Given a decision-making state $s^i$, an action $a \in A$ is taken.

\item This action results in the system receiving an instantaneous reward (or incurring a cost) which is defined by the function $\rho(s^i,a)$.

\item Upon taking this action, the system transitions to state $s^j$ according to the probability distribution $p_{ij}(a)$

\item The transitions are however, not instantaneous. The system takes time $t$ to make the transition, where $t \sim t_{ij}$.
\end{enumerate}

\begin{table}[h]
% \small
\caption{Algorithm Hyper-parameter Description}
\footnotesize
\begin{tabular}{>{\raggedright\arraybackslash}p{2.1cm} p{5.0cm}}
\toprule

\textbf{Candidate Action Factor $\epsilon$: } &
    $\epsilon$ controls the number of responders considered for each incident: any responder with response time within $\epsilon$ times the greedy action's response time is considered. Therefore $\epsilon$ directly controls the branching factor of the MCTS tree, and has the most significant impact on computation time.  \\ 
    \hline

\textbf{Simulation Budget $b$: }  &
    $b$ is the number of incident chains that are generated and evaluated from the incident model $\Theta$. Increasing $b$ decreases the variance inherent when sampling. Each chain of incidents can be processed in parallel, so increasing $b$ does not directly increase computation time, assuming enough computing resources are available.  
    \\
    \hline

\textbf{Stochastic Horizon $h^{s}$: } &
    $h^{s}$ is the number of predicted incidents to explore in the future before defaulting to the greedy action. Increasing $h^{s}$ also has a pronounced affect on computation time. Each level of the search tree that is expanded increases the number of states to simulate. 
    \\
    \hline

\textbf{Discount \t Factor $\gamma$: }  &
    $\gamma$ is the discount applied to future predicted incident rewards, and is in the range (0,1). High values of $\gamma$ weight far future incidents more similarly to those about to happen, while low values give much more weight to incidents that are happening soon. Unlike the other hyper-parameters, varying $\gamma$ has no effect on computation time, a value that minimizes response times for each environment should be chosen. 
    \\
    \bottomrule
\end{tabular}
% \label{tbl:hyperparam_desc}
% \vspace{-0.2in}
\end{table}

\begin{table}[h]
 \small
 \renewcommand{\arraystretch}{1.5}
 \centering
 \caption{Summary and dimension of implemented features for traffic prediction model.}
 \vspace{0.05in}
 \label{table:features_table}
 \begin{tabular}{>{\raggedright\arraybackslash}p{2.1cm} p{0.6cm} p{4.0cm}}
 \toprule
 \textbf{Feature} & \textbf{Dim.} & \textbf{Description}\\ 
 \midrule
 Hour of day, Day of week & 2 & Hour of the day and Day of week used to sample speed data\\
 Length & 1 & Length of the street segment, collected from OpenStreetMap\\
 Freeflow speed & 1 & Freeflow speed on the street segment, collected from HERE API. This dataset is private and is collected by our research group \\
 Number of lanes & 1 & Number of lanes on the street segment, collected from OpenStreetMap \\
 TAZ & 741 & Binary indication of Traffic Analysis Zone (TAZ) corresponding to this feature vector, collected from US Census Bureau. A TAZ can contain multiple network segments. \\
 Realtime speed & 1 & The realtime speed value collected from HERE API. This dataset is private and is collected by our research group \\
 \bottomrule
 \end{tabular}
 \end{table}

The first step (and the most general approach) in solving such a problem is to convert this into a Discrete-Time Markovian process. However, even after discretization, it is particularly challenging to solve this problem since - a) the state-space is practically intractable, and b) the state transition probabilities $p_{ij}(a)$ are unknown. In order to alleviate these issues, \textit{SimTrans} uses canonical policy iteration with an added computational step. At each step of policy iteration, it uses a simulator to estimate values of states it encounters; this provides crucial data about how state transitions occur in the system, which is then used to learn the distribution $p_{ij}(a)$.

While policy iteration is guaranteed to converge to the optimal policy, this approach has major drawbacks. First, due to the size of the state-space, even on fairly powerful computing systems, it takes weeks to learn the optimal policy. Finding such a policy with the inclusion of environment variables would be computationally infeasible. Secondly, and partly as a consequence of the first issue, \textit{SimTrans} is simply not suited for for dynamic environments: if a single responder breaks down, traffic conditions change, or incident models evolve, the policy must be relearned from scratch, which takes time that is incomparable to the latency that such emergency responder systems can afford.

In order to alleviate these concerns, the paper described an algorithm based on Monte-Carlo Tree Search, that looks to learn a near-optimal action for a given state only, instead of focussing on learning a policy over the entire state space. We describe the algorithm in the main paper, but present a summary of the exogenous parameters here in the appendix, for easy reference.

%\subsection{Monte Carlo Tree Search}
%
%Monte Carlo Tree Search (MCTS) is a class of heuristic-based search algorithms to compute the most promising actions for a given state of a decision-making problem. Given an arbitrary state of the problem, the essential focus of the MCTS is to repeatedly simulate the system under consideration to build a search tree that keeps track of the actions taken and the rewards received. The search tree is built in a manner that balances exploration (trying new or sub-optimal actions) and exploitation (greedily exploring the best action). The simulations are then used to make a decision for the root state. 

\balance

\subsection{Traffic and Congestion Prediction}

As mentioned in the main paper, we assume that the entire urban area under consideration is divided into a set of road segments $V$ and every $v_i \in V$ has a set of features associated with it.

For building the search tree for the dispatching approach described in the section \ref{sec:dispatchAlg}, we need to accurately estimate the time it takes for a responder to get to an incident. In order to facilitate this, we developed a predictive model to estimate speed on each road segment for a given time interval. The model needs to be contextualized with features that affect speed, and from our experience and prior work, we chose hour of day, day of week, number of lanes on the road, and the traffic analysis zone as our principal features.  These features are described in Table \ref{table:features_table}.

To find the best optimizer for our LSTM model, we trained it with three different optimizers-- Adam \cite{kingma2014adam}, SGD \cite{robbins1985stochastic}, Adagrad \cite{duchi2011adaptive} and model performance was evaluated using five-fold shuffled cross validation. Table \ref{table:lstm_tuning_table} shows Mean absolute error (MAE) in miles/hour units for different optimizers. The result shows that SGD performs better with our LSTM model than other optimizers, hence it is chosen for our route finding algorithm. In test dataset, LSTM model with SGD optimizer has MAE of \textbf{6.419 miles/hour}. There are other state-of-the-art models for estimating traffic speeds \cite{yu2017spatiotemporal, li2018diffusion}, however, we have not explored them in this paper. Our system design is modular and other algorithms can easily fit in.

\begin{table}
\renewcommand{\arraystretch}{1.5}
\centering
\caption{LSTM Hyperparameter tuning table}
\vspace{0.05in}
\label{table:lstm_tuning_table}
\begin{tabular}{>{\raggedright\arraybackslash}p{2.5cm} p{1.5cm}}
\toprule
\textbf{Optimizer} & \textbf{MAE}\\ 
\midrule
Adam & 5.47\\
SGD & 4.27\\
Adagrad & 6.16\\
\bottomrule\\
\end{tabular}
\end{table}

\subsection{Real Incident Response Times}

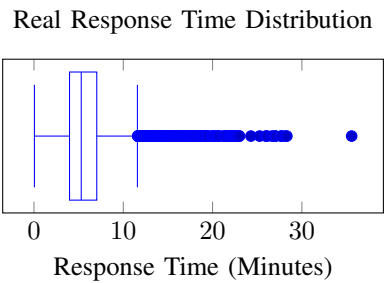
\begin{figure}
    \centering
\begin{tikzpicture}
\begin{axis}
[
      width=.75\columnwidth,
      height=0.20\textwidth,
    %   font=\footnotesize,
    %   grid=major,
    %   xtick=,
    ymajorticks=false,
      title={Real Response Time Distribution},
      xlabel={Response Time (Minutes)},
    %   ylabel near ticks
    ]
      \addplot+[boxplot={draw position=1}] table [col sep=comma, y expr=\thisrow{time}*(1/60)] {data/resp_time_data/real_incident_resp.csv};
      
\end{axis}
\end{tikzpicture}

\caption{Response time distribution for EMS response for the Nashville Fire Department from 01-02-2016 to 01-02-2017.
%\textcolor{red}{fill caption}
% Results from Experiment 2.   The horizontal axis is the elapsed time in seconds. The values shown from bottom to top are: (a) time to register job offer, (b) time to register resource offer, (c) time to match offers (measured from when offers were issued), (d) time to register all actors, (e) time to select mediators before posting job offer, (f) time to complete job (measured from when the job was first posted), and (g) the execution time of the job.
}
\label{fig:real_resp_boxplot}
\end{figure}

Figure \ref{fig:real_resp_boxplot} shows the Nashville Fire Department's actual response times to incidents for one year of data from February 2016 to February 2017.

%This is Empty File

\bibliographystyle{IEEEtran}
\bibliography{paper}
\end{document}